\documentclass{article}

\usepackage{microtype}
\usepackage{graphicx}
\usepackage{subfigure}
\usepackage{booktabs} % for professional tables

\usepackage[colorinlistoftodos,bordercolor=orange,backgroundcolor=orange!20,linecolor=orange,textsize=scriptsize]{todonotes}

\usepackage{hyperref}

\usepackage[accepted]{icml2019blank}

\icmltitlerunning{SAGA with Arbitrary Sampling}

\usepackage{amsmath,amsfonts,amssymb,amsthm,array}

\usepackage{mdframed} 
\usepackage{thmtools}

\newcommand{\Prob}{\mathbb{P}}

\newcommand{\R}{\mathbb{R}}
\newcommand{\Exp}{\mathbb{E}}
\def\<#1,#2>{\langle #1,#2\rangle}
\newcommand{\cX}{\mathcal{X}}
\newcommand{\bJ}{\mathbf{J}}
\newcommand{\dom}{\operatorname{dom}}
\newcommand{\prox}{\operatorname{prox}}

\newcommand{\eqdef}{\; { := }\;}

\definecolor{shadecolor}{gray}{0.90}
\declaretheoremstyle[
headfont=\normalfont\bfseries,
notefont=\mdseries, notebraces={(}{)},
bodyfont=\normalfont,
postheadspace=0.5em,
spaceabove=1pt,
mdframed={
	skipabove=8pt,
	skipbelow=8pt,
	hidealllines=true,
	backgroundcolor={shadecolor},
	innerleftmargin=4pt,
	innerrightmargin=4pt}
]{shaded}

\declaretheorem[style=shaded,within=section]{definition}
\declaretheorem[style=shaded,sibling=definition]{theorem}

\declaretheorem[style=shaded,sibling=definition]{assumption}

\declaretheorem[style=shaded,sibling=definition]{lemma}

\declaretheorem[style=shaded,sibling=definition]{remark}

\begin{document}

\twocolumn[
%\icmltitle{Minibath SAGA for Arbitrary Sampling}
\icmltitle{SAGA with Arbitrary Sampling}




\icmlsetsymbol{equal}{*}

\begin{icmlauthorlist}
\icmlauthor{Xun Qian}{kaust}
\icmlauthor{Zheng Qu}{hk}
\icmlauthor{Peter Richt\'{a}rik}{kaust,edi,mipt}

\end{icmlauthorlist}

\icmlaffiliation{kaust}{King Abdullah University of Science and Technology, Kingdom of Saudi Arabia}
\icmlaffiliation{mipt}{Moscow Institute of Physics and Technology, Russian Federation}
\icmlaffiliation{edi}{University of Edinburgh, United Kingdom}
\icmlaffiliation{hk}{The University of Hong Kong, Hong Kong}

\icmlcorrespondingauthor{Xun Qian}{xun.qian@kaust.edu.sa}
\icmlcorrespondingauthor{Zheng Qu}{zhengqu@hku.hk}
\icmlcorrespondingauthor{Peter Richt\'{a}rik}{peter.richtarik@kaust.edu.sa}

\icmlkeywords{SAGA, arbitrary sampling, SGD, variance reduction, importance sampling}

\vskip 0.3in
]



\printAffiliationsAndNotice{}  % leave blank if no need to mention equal contribution

\begin{abstract}
We study the problem of minimizing the average of a very large number of smooth functions, which is of key importance in  training supervised  learning models.  One of the most celebrated methods in this context is the SAGA algorithm of \citet{SAGA}. Despite years of research on the topic, a general-purpose version of SAGA---one that would include arbitrary importance sampling and minibatching schemes---does not exist.  We remedy this situation and propose a general and flexible variant of SAGA following the {\em arbitrary sampling}  paradigm. We perform an iteration complexity analysis of the method, largely possible due to the construction of  new stochastic Lyapunov functions. We establish linear convergence rates in the smooth and  strongly convex regime, and  under a quadratic functional growth condition (i.e.,  in a  regime not assuming strong convexity). 
% SAGA is the only variance-reduced method achievinbg linear convergence without any a-priori knowldge of the error bound condition number.  
Our rates match those of the primal-dual method Quartz~\cite{Quartz} for which an arbitrary sampling analysis is available, which makes a significant step towards closing the gap in our understanding of complexity of primal and dual methods for finite sum problems. 
% Finally, we show through experiments that specific variants of our general SAGA method can perform better in practice than other competing methods.
\end{abstract}

\section{Introduction}

We consider a convex composite optimization problem
\begin{equation}\label{primal}
\textstyle \min_{x\in \mathbb{R}^d} P(x) \eqdef \left( \sum\limits_{i=1}^n \lambda_i f_i(x)\right) + \psi(x),
\end{equation}
where $f:=\sum_i \lambda_i f_i$ is a conic combination (with coefficients $\lambda_1,\dots,\lambda_n > 0$) of a very large number of smooth convex functions  $f_i:\R^d\to \R$, and   $\psi:\R^d \to \R\cup \{+\infty\}$ is a proper closed convex function.\footnote{Our choice to consider general weights $\lambda_i>$ is not significant; we do it for convenience as will be clear from the relevant part of the text. Hence, the reader may without the risk of missing any key ideas assume that $\lambda_i=\tfrac{1}{n}$ for all $i$.}  We do not assume $\psi$ to be smooth. In particular, $\psi$ can be the indicator function of a nonempty closed convex set, turning problem \eqref{primal} into a constrained minimization of function $f$. We are  interested in the regime where $n\gg d$, although all our theoretical results hold without this assumption.

In a typical setup in the literature, $\lambda_i=1/n$ for all $i\in [n]\eqdef \{1,2,\dots,n\}$, $f_i(x)$ corresponds to the loss of a supervised machine learning model $x$ on example $i$ from a training dataset  of size $n$, and $f$ represents the average loss (i.e., empirical risk). Problems of the form \eqref{primal} are often called ``finite-sum'' or regularized empirical risk minimization (ERM) problems, and are of immense importance in supervised learning, essentially forming the dominant training paradigm~\cite{shai_book-2014}.

 \subsection{Variance-reduced methods}
 
One of the most successful methods for solving ERM problems is stochastic gradient descent (SGD) \cite{RobbinsMonro:1951,  Nemirovski-Juditsky-Lan-Shapiro-2009}  and its many variants, including those with minibatches~\cite{pegasos2}, importance sampling~\cite{NeedellWard2015, IProx-SDCA} and momentum~\cite{SHB-NIPS, SMOMENTUM}. 

One of the most interesting developments in recent years concerns {\em variance-reduced} variants of SGD. The first method in this category is the celebrated\footnote{\citet{schmidt2017minimizing} received the 2018 Lagrange Prize in continuous optimization for their work on SAG.} stochastic average gradient (SAG) method of \citet{schmidt2017minimizing}. Many additional variance-reduced methods were proposed since, including SDCA~\cite{UCDC, SDCA, SDCA-dual-free}, SAGA~\cite{SAGA}, SVRG~\cite{johnson2013accelerating, proxSVRG}, S2GD~\cite{S2GD, mS2GD}, MISO~\cite{MISO}, JacSketch~\cite{Jacsketch} and SAGD~\cite{SAGD}. 

% SARAH~\cite{SARAH}, 
% dfSDCA

\subsection{SAGA: the known and the unknown}

Since the SAG gradient estimator is not  unbiased, SAG is notoriously hard to analyze. Soon after SAG was proposed, the SAGA method \cite{SAGA} was developed, obtained by replacing the biased SAG estimator by a similar, but unbiased, SAGA estimator. This method admits a simpler analysis, retaining the favourable convergence properties of SAG. SAGA is one of the early and most successful variance-reduced methods for \eqref{primal}. 

Better understanding of the behaviour of SAGA remains one of the open challenges in the literature. Consider problem \eqref{primal} with $\lambda_i=1/n$ for all $i$. Assume, for simplicity, that each $f_i$ is $L_i$-smooth and $f$ is $\mu$-strongly convex. In this regime, the iteration complexity of SAGA with uniform sampling probabilities  is ${\cal O}((n + \tfrac{L_{\max}}{\mu})\log\tfrac{1}{\epsilon})$, where $L_{\max}\eqdef \max_i L_i$, which was  established already by~\citet{SAGA}. \citet{mschmidt-aistats2015} conjectured that there exist nonuniform sampling probabilities for which the complexity improves to ${\cal O}((n+\tfrac{\bar L}{\mu})\log\tfrac{1}{\epsilon})$, where ${\bar L} \eqdef \sum_iL_i/n$. However, the ``correct'' importance sampling strategy  leading to this result  was not discovered until recently in the work of \citet{Jacsketch}, where the conjecture was resolved in the affirmative.  One of the key difficulties in the analysis was the construction of a suitable stochastic Lyapunov function controlling the iterative process. Likewise, until recently, very little was known about the minibatch performance of SAGA, even for the simplest {\em uniform} minibatch strategies. Notable advancements in this area were made by \citet{Jacsketch}, who have the currently best rates for SAGA with standard uniform minibatch strategies and the first importance sampling results for a block variant of SAGA.

\subsection{Contributions}

\begin{table*}[t]
\begin{center}
{\footnotesize
\begin{tabular}{|c|c|c|}
\hline
%\abovespace\belowspace
 Regime & Arbitrary sampling &    Theorem \\
\hline 
% & &  \\
\begin{tabular}{c} {\bf Smooth}\\
$\psi\equiv 0$  \\
$f_i$ is $L_i$-smooth, $f$ is $\mu$-strongly convex
\end{tabular}  & 
$ \displaystyle \max_{1\leq i \leq n}\left\{ \tfrac{1}{{\color{red}p_i}} + \tfrac{4L_i {\color{blue}\beta_i}\lambda_i}{\mu} \right\}  \log \left(\tfrac{1}{\epsilon}\right) $ & \ref{Th:convp} \\
% & &  \\
\hline
\begin{tabular}{c} 
{\bf Nonsmooth} \\
$P$ satisfies $\mu$-growth condition \eqref{eq:gc} and Assumption~\ref{ass:uniquegradient} \\
$f_i(x) = \phi_i({\bf A}_i^\top x)$, $\phi_i$ is $1/\gamma$-smooth, $f$ is $L$-smooth
\end{tabular} & 
$\displaystyle \left(2 + \max\left\{ \tfrac{6L}{\mu}, 3\max_{1\leq i\leq n} \left\{ \tfrac{1}{{\color{red}p_i}} + \tfrac{4 {\color{blue}v_i} \lambda_i}{ {\color{red}p_i} \mu\gamma} \right\}  \right\} \right) \log \left(\tfrac{1}{\epsilon}\right)$   & \ref{th:phi}     \\ 
%& &  \\
\hline
\begin{tabular}{c} 
{\bf Nonsmooth} \\
$\psi$ is $\mu$-strongly convex\\
$f_i(x) = \phi_i({\bf A}_i^\top x)$, $\phi_i$ is $1/\gamma$-smooth
\end{tabular} 
& $\displaystyle \max_{1\leq i\leq n} \left\{ 1+ \tfrac{1}{{\color{red}p_i}} + \tfrac{3{\color{blue}v_i}\lambda_i}{{\color{red}p_i}\mu \gamma} \right\} \log \left(\tfrac{1}{\epsilon}\right) $   & \ref{th:phistrongly}   \\ 
%& &  \\
%\abovespace\belowspace
\hline
\end{tabular}
}
\end{center} 
\caption{Iteration complexity results for SAGA-AS. We have ${\color{red} p_i}\eqdef \Prob(i\in S)$, where $S$ is a  sampling of subsets of $[n]$ utilized by SAGA-AS. The key complexity parameters $\color{blue}\beta_i$ and $\color{blue} v_i$ are defined in the sections containing the theorems.}
\label{tab:summary}
\end{table*}

Our contributions can be summarized as follows.

 {\bf SAGA with arbitrary sampling.} We study the performance of SAGA under fully general data sampling strategies known in the literature as {\em arbitrary sampling}, generalizing all previous results, and obtaining an array of new theoretically and practically useful samplings. We call our general method SAGA-AS. Our theorems are expressed in terms of new stochastic and deterministic Lyapunov functions, the constructions of which was essential to our success.
 
 In the arbitrary sampling paradigm, first proposed by \citet{NSYNC} in the context of randomized coordinate descent methods, one  considers all (proper) random set valued mappings $S$ (called ``samplings'') with values being subsets of $[n]$. A sampling is uniquely determined by assigning a probability to all $2^n$ subsets of $[n]$. A sampling is called proper\footnote{It does not make sense to consider samplings $S$ that are not proper. Indeed, if $p_i=0$ for some $i$, a method based on $S$ will lose access to $f_i$ and, consequently,  ability to  solve \eqref{primal}.} if probability of each $i\in [n]$ being sampled is positive; that is, if $p_i\eqdef \Prob(i\in S)>0$ for all $i$. Finally, the term  ``arbitrary sampling''  refers to an arbitrary proper sampling.

{\bf Smooth case.} We perform an iteration complexity analysis in the smooth case ($\psi \equiv 0$), assuming $f$ is $\mu$-strongly convex. Our analysis generalizes the results of \citet{SAGA} and \citet{Jacsketch} to arbitrary sampling. The JacSketch method~\citet{Jacsketch} and its analysis rely on the notion of a bias-correcting random variable. Unfortunately, such a random variable does not exist for SAGA-AS. We overcome this obstacle by proposing a {\em bias-correcting  random vector (BCRV)}  which, as we show, always exists. While \citet{Jacsketch, SAGD} consider particular suboptimal choices, {\em we are able to find the BCRV which minimizes the iteration complexity bound}. Unlike all known and new variants of SAGA considered in \cite{Jacsketch}, {\em SAGA-AS does not arise as a special case of JacSketch}.  Our linear rates for SAGA-AS  are the same as those for the primal-dual stochastic fixed point method Quartz~\cite{Quartz} (the first arbitrary sampling based method for \eqref{primal}) in the regime when Quartz is applicable, which is the case when an explicit strongly convex regularizer is present. In contrast, we do not need an explicit regularizer, which means that {\em SAGA-AS can utilize the strong convexity of $f$ fully}, even if the strong convexity parameter $\mu$ is not known. While the importance sampling results in \cite{Jacsketch} require each $f_i$ to be strongly convex, we impose this requirement on $f$ only.

 {\bf Nonsmooth case.} We perform an iteration complexity analysis in the general nonsmooth case. When the regularizer $\psi$ is strongly convex, which is the same setting as that considered in \cite{Quartz}, our iteration complexity bounds are essentially the same as that of Quartz. However,  we also prove linear convergence results, with the same rates, under a quadratic functional growth condition (which does not imply strong convexity) \cite{Necoara-Nesterov-Glineur-2018-linear-without-strong-convexity}. These are {\em the first linear convergence result for any variant of SAGA without strong convexity.} Moreover, to the best of our knowledge, {\em SAGA-AS is the only variance-reduced method which achieves linear convergence without any a priori knowledge of the error bound condition number.}

Our arbitrary sampling rates are summarized in Table~\ref{tab:summary}.

\subsection{Brief review of arbitrary sampling results}

The arbitrary sampling paradigm was proposed by \citet{NSYNC}, where a randomized coordinate descent method with arbitrary sampling of subsets of coordinates was analyzed for unconstrained minimization of a strongly convex function. Subsequently, the primal-dual method  Quartz with arbitrary sampling of dual variables was studied in \citet{Quartz} for solving \eqref{primal} in the case when $\psi$ is strongly convex (and $\lambda_i=\tfrac{1}{n}$ for all $i$). An accelerated randomized coordinate descent method with arbitrary sampling called ALPHA was proposed  by \citet{qu2016coordinate} in the context of  minimizing the sum of a smooth convex function and a convex block-separable regularizer. A key concept in the analysis of all known methods in the arbitrary sampling paradigm is the notion of {\em expected separable overapproximation (ESO)}, introduced by \citet{PCDM}, and in the context of arbitrary sampling studied in depth by \citet{qu2016coordinate2}. A stochastic primal-dual hybrid gradient algorithm (aka Chambolle-Pock) with arbitrary sampling of dual variables was studied by \citet{SCP17}. Recently, an accelerated coordinate descent method with arbitrary sampling  for minimizing a smooth and strongly convex function was studied by \citet{FilipACD18}. Finally, the first arbitrary sampling analysis in a nonconvex setting was performed by \citet{nonconvex_arbitrary}, which is also the first work in which the optimal sampling out of class of all samplings of a given minibatch size was identified.

\section{The Algorithm}

Let $F: \mathbb{R}^d \to \mathbb{R}^n$ be defined by
$$
\textstyle F(x) \eqdef (f_1(x), ..., f_n(x))^\top$$
and let 
$
 {\bf G}(x) \eqdef [\nabla f_1(x), ..., \nabla f_n(x)] \in {\mathbb{R}^{d\times n}}
$
be the Jacobian of $F$ at $x$. 

\subsection{JacSketch}

\citet{Jacsketch} propose a new  family of variance reduced SGD methods---called JacSketch---which progressively build a variance-reduced estimator of the gradient via the utilization of a new technique they call {\em Jacobian sketching}. As shown in \cite{Jacsketch}, state-of-the-art variants SAGA can be obtained as a special case of JacSketch. However, {\em SAGA-AS does not arise as a special case of  JacSketch}. In fact, the generic analysis provided in \cite{Jacsketch} (Theorem~3.6) is too coarse and does not lead to good bounds for any variants of SAGA with importance sampling. On the other hand, the analysis of \citet{Jacsketch} which does do well for importance sampling does not generalize to arbitrary sampling, regularized objectives or regimes without strong convexity. 

In this section we provide a brief review of the JacSketch method, establishing some useful notation along the way, with the goal of pointing out the moment of departure from JacSketch construction which  leads to SAGA-AS.
The iterations of JacSketch have the form 
\begin{equation}\label{eq:JS}
x^{k+1} = x^k - \alpha g^k,
\end{equation}
where $\alpha>0$ is a fixed step size, and $g^k$ is an variance-reduced unbiased estimator of the gradient $\nabla f(x^k)$ built iteratively through a process involving Jacobian sketching, sketch-and-project and bias-correction.

Starting from an arbitrary matrix ${\bf J}^0\in \R^{d\times n}$, at each $k\geq 0$ of JacSketch an estimator  ${\bf J}^k \in \R^{d\times n}$ of the true Jacobian  ${\bf G}(x^k)$ is constructed using a sketch-and-project iteration:
\begin{equation} \label{eq:SandP}
\begin{array}{rcl}
{\bf J}^{k+1} = & & \arg \min\limits_{{\bf J}\in \mathbb{R}^{d\times n} } \| {\bf J} - {\bf J}^k \| \\
& & {\rm subject \ to} \ {\bf J}{\bf S}_k =  {\bf G}(x^k){\bf S}_k.
\end{array}
\end{equation}
Above,  $\|{\bf X}\| \eqdef \sqrt{{\rm Tr}({\bf X}{\bf X}^\top)}$ is the Frobenius norm\footnote{\citet{Jacsketch} consider a {\em weighted} Frobenius norm, but this is not useful for our purposes.}, and ${\bf S}_k \in {\mathbb{R}}^{n\times \tau}$  is a random matrix drawn from some ensamble of matrices $\cal D$ in an i.i.d.\ fashion in each iteration. The solution to \eqref{eq:SandP} is the closest matrix to ${\bf J}^k$ consistent with  true Jacobian in its action onto ${\bf S}_k$. Intuitively, the higher $\tau$ is, the more accurate ${\bf J}^{k+1}$ will be as an estimator of the true Jacobian ${\bf G}(x^k)$. However, in order to control the cost of computing ${\bf J}^{k+1}$, in practical applications one chooses   $\tau\ll n$. In the case of standard SAGA, for instance, ${\bf S}_k$ is a random standard unit basis vector in $\R^n$ chosen uniformly at random (and hence $\tau=1$).

The projection subproblem \eqref{eq:SandP} has the explicit solution (see Lemma~B.1 in \cite{Jacsketch}): 
$$
{\bf J}^{k+1} = {\bf J}^k + ( {\bf G}(x^k) - {\bf J}^k){\bf \Pi}_{{\bf S}_k},
$$
where 
$
{\bf \Pi }_{\bf S} \eqdef {\bf S}({\bf S}^\top {\bf S})^{\dagger}{\bf S}^\top
$
is an orthogonal projection matrix (onto the column space of $\bf S$), and $\dagger$ denotes the Moore-Penrose pseudoinverse. Since ${\bf J}^{k+1}$ is constructed to be an approximation of ${\bf G}(x^k)$, and since $\nabla f(x^k) = {\bf G}(x^k) \lambda$, where $\lambda \eqdef (\lambda_1, ..., \lambda_n)^\top$, it makes sense to estimate the gradient via $\nabla f(x^k)\approx {\bf J}^{k+1} \lambda$.
However, this gradient estimator is not unbiased, which poses dramatic challenges for complexity analysis. Indeed, the celebrated SAG method, with its infamously technical analysis, uses precisely this estimator in the special case when ${\bf S}_k$ is chosen to be a standard basis vector in $\R^n$ sampled uniformly at random. Fortunately, as show in \cite{Jacsketch}, an unbiased estimator can be constructed by taking a random linear combination of ${\bf J}^{k+1} \lambda$ and ${\bf J}^{k} \lambda$:
\begin{eqnarray}
g^k &=& (1-\theta_{{\bf S}_k}){\bf J}^{k} \lambda + \theta_{{\bf S}_k} {\bf J}^{k+1} \lambda \notag \\
&=& {\bf J}^k \lambda + \theta_{{\bf S}_k}({\bf G}(x^k) - {\bf J}^k){\bf \Pi}_{{\bf S}_k}\lambda, \label{eq:b9g8f9ff}
\end{eqnarray} 
where $\theta = \theta_{{\bf S}}$ is a bias-correcting random variable (dependent on $\bf S$), defined as any random variable for which $\Exp[\theta_{\bf S} {\bf \Pi}_{\bf S} \lambda] = \lambda.$ Under this condition, $g^k$ becomes an unbiased estimator of $\nabla f(x^k)$.  The JacSketch method is obtained by alternating  optimization steps \eqref{eq:JS} (producing iterates $x^k$) with sketch-and-project steps (producing  ${\bf J}^k$).

\subsection{Bias-correcting random vector}

% \xun{Maybe comment on this, maybe not. Noticing that $\mathbb{E}[\theta_{S}{\bf \Pi}_{\bf S} e] = e$ may not be guaranteed for arbitrary sampling (for instance, the sampling: $\mathbb{P}[\{1,2\}] = \mathbb{P}[\{2, 3\}] = \tfrac{1}{2}$ with $n=3$), we use a random diagonal matrix $ \theta_{S}$ instead. 
% }
% 
In order to construct SAGA-LS, we take a departure here and consider a {\em bias-correcting random vector} \[\theta_{\bf S} = (\theta_{\bf S}^1,\cdots, \theta_{\bf S}^n)^\top \in \R^n\] instead. From now on it will be useful to think of $\theta_{\bf S}$ as an $n\times n$ diagonal matrix, with the vector embedded in its diagonal. In contrast to \eqref{eq:b9g8f9ff}, we propose to construct $g^k$ via
\[g^k =  {\bf J}^k \lambda + ({\bf G}(x^k) - {\bf J}^k) \theta_{{\bf S}_k} {\bf \Pi}_{{\bf S}_k}\lambda.\]
It is easy to see that under the following assumption, $g^k$ will be an unbiased estimator of $\nabla f(x^k)$.

\begin{assumption}[Bias-correcting random vector]\label{ass:thetaS}
We say that the diagonal random matrix $\theta_{\bf S} \in \mathbb{R}^{n\times n}$ is a bias-correcting random vector if
\begin{equation}
\mathbb{E}[\theta_{\bf S}{\bf \Pi}_{\bf S} \lambda] = \lambda.\label{eq:BC-cond}
\end{equation}
\end{assumption}

\subsection{Choosing  distribution $\cal D$}

In order to complete the description of SAGA-AS, we need to specify the distribution $\cal D$. We choose $\cal D$ to be a distribution over random column submatrices of the $n\times n$ identity matrix $\bf I$. Such a distribution is uniquely characterized by a random subset of the columns of ${\bf I}\in \R^{n\times n}$, i.e., a random subset of $[n]$. This leads us to the notion of  a {\em sampling}, already outlined in the introduction.

\begin{definition}[Sampling] 
A {\em sampling} $S$ is a random set-valued mapping with values being the subsets of $[n]$. It is uniquely characterized by the choice of probabilities $p_C \eqdef \Prob[S=C]$ associated with every subset $C$ of $[n]$.
Given a sampling $S$, we let $p_i \eqdef \Prob[i \in S] = \sum_{C:i\in C}p_C$. We say that $S$ is {\em proper} if $p_i>0$ for all $i$. 
\end{definition}

So, given a proper sampling $S$, we sample matrices $\bf S\sim {\cal D}$ as follows: i) Draw a random set $S$, ii) Define ${\bf S} = {\bf I}_{:S}\in \R^{n\times |S|}$ (random column submatrix of $\bf I$ corresponding to columns $i\in S$). For $h = (h_1,\cdots,h_n)^\top \in \R^n$ and sampling $S$ define vector $h_S\in \R^n$ as follows:
$$
(h_{S})_i \eqdef h_i 1_{(i\in S)}, \; \text{where} \; 1_{(i\in S)} \eqdef \left\{ \begin{array}{rl}
1, & \mbox{ if $i \in { S}$} \\
0, &\mbox{ otherwise.}
\end{array} \right.
$$

It is easy to observe (see Lemma 4.7 in \cite{Jacsketch}) that for ${\bf S} = {\bf I}_{:S}$ we have the identity
\begin{equation}\label{eq:nb98fg98bf}{\bf \Pi }_{\bf S} = {\bf \Pi }_{{\bf I}_{:S}}  = {\bf I}_S \eqdef {\rm Diag}(e_S).\end{equation}
In order to simplify notation, we will write $\theta_S$ instead of $\theta_{\bf S} = \theta_{{\bf I}_{:S}}$. 

\begin{lemma} \label{lem:n98hf9hf} Let $S$ be a proper sampling and define $\cal D$ by setting ${\bf S} = {\bf I}_{:S}$. Then 
 condition~\eqref{eq:BC-cond} is equivalent to
\begin{equation} \label{eq:nbdfg98ud8df}\textstyle \Exp[ \theta_S^i 1_{(i\in S)}] \equiv \sum_{C \subseteq [n] : i\in C} p_C \theta_C^i  = 1, \quad \forall i\in [
n].\end{equation}
This condition is satisfied by the \em{default} vector $\theta_{S}^i \equiv \tfrac{1}{p_i}$.
\end{lemma}

In general, there is an infinity of bias-correcting random vectors characterized by \eqref{eq:nbdfg98ud8df}. In SAGA-AS we reserve the freedom to choose any of these vectors.

\subsection{SAGA-AS}

By putting all of the development above together, we have arrived at SAGA-AS (Algorithm \ref{minibatchSAGA}). Note that since we consider problem \eqref{primal} with a regularizer $\psi$, the optimization step involves a proximal operator, defined as
\[\prox^\psi_{\alpha}(x)\eqdef \arg\min \left\{\tfrac{1}{2\alpha }\| x-y\|^2+\psi(y) \right\}, \quad \alpha>0.\]
To shed more light onto the key steps of SAGA-AS, note that that an alternative way of writing the Jacobian update  is
\[{\bf J}^{k+1}_{:i} = \begin{cases} {\bf J}^{k}_{:i} & \quad \text{if} \quad i\notin S_k,\\
\nabla f_i(x^k) & \quad \text{if} \quad i\in S_k, \end{cases}\]
while the gradient estimate can be alternatively written as
\begin{align*}g^k &=\textstyle  \sum_{i=1}^n \lambda_i {\bf J}^k_{:i} +  \sum_{i\in S_k}\lambda_i \theta_{S_k}^i \left[ \nabla f_i(x^k) - {\bf J}^{k}_{:i} \right] \\
& \textstyle = \sum_{i \notin S_k} \lambda_i {\bf J}^k_{:i} + \sum_{i\in S_k} \lambda_i  \left[\theta_{S_k}^i \nabla f_i(x^k) + (1-\theta_{S_k}^i ) {\bf J}^k_{:i} \right]. 
\end{align*}

\begin{algorithm}[tb]
	%\caption{Minibatch SAGA with Arbitrary Sampling}
	\caption{SAGA with Arbitrary Sampling (SAGA-AS)}
	\label{minibatchSAGA}
	\begin{algorithmic}
		\STATE {\bfseries Parameters:} Arbitrary proper sampling $S$; bias-correcting random vector $\theta_S$; stepsize $\alpha > 0$
		\STATE {\bfseries Initialization:} Choose $x^0 \in \mathbb{R}^d$, ${\bf J}^0 \in \mathbb{R}^{d\times n}$ 
		\FOR{ $k = 0, 1, 2, ...$}
		\STATE Sample a fresh set $S_k \sim S\subseteq [n]$ 
%		\STATE ${\bf J}^{k+1} = {\bf J}^k + ({\bf G}(x^k) - {\bf J}^k){\bf \Pi}_{{\bf I}_{:S_k}}$ 
\STATE ${\bf J}^{k+1} = {\bf J}^k + ({\bf G}(x^k) - {\bf J}^k)  {\bf I}_{S_k}$ 
%		\STATE $g^k = {\bf J}^k\lambda + ( {\bf G}(x^k) - {\bf J}^k){\theta_{S_k}} {\bf \Pi}_{{\bf I}_{:S_k}} \lambda$ 
\STATE $g^k = {\bf J}^k\lambda + ( {\bf G}(x^k) - {\bf J}^k){\theta_{S_k}} {\bf I}_{S_k} \lambda$ 
%		\STATE $x^{k+1} = x^k -\alpha g^k$ 
		\STATE $x^{k+1} =\prox^{\psi}_{\alpha}\left( x^k -\alpha g^k\right)$ 		
		\ENDFOR
	\end{algorithmic}
\end{algorithm}

\section{Analysis in the Smooth Case} \label{sec:smooth}

In this section we consider problem \eqref{primal} in the smooth case; i.e., we let $\psi\equiv 0$. 

\subsection{Main result} Given an arbitrary proper sampling $S$, and bias-correcting random vector $\theta_S$, for each $i\in [n]$ define 
\begin{equation}\label{betai}
\textstyle \beta_i \eqdef \sum_{C \subseteq[n] : i\in C}p_C|C|(\theta_C^i)^2, \quad i\in [n],
\end{equation}
where $|C|$ is the cardinality of the set $C$. As we shall see, these  quantities play a key importance in our complexity result, presented next.

\begin{theorem}\label{Th:convp}
Let $S$ be an arbitrary proper sampling, and let ${\theta}_S$ be a bias-correcting random vector satisfying \eqref{eq:nbdfg98ud8df}. Let $f$ be $\mu$-strongly convex and $f_i$ be convex and $L_i$-smooth. 
Let $\{ x^k, {\bf J}^k \}$ be the iterates produced by Algorithm \ref{minibatchSAGA}. Consider the stochastic Lyapunov function 
$$
\Psi^k_S \eqdef \|x^k -x^*\|^2 + 2\alpha |S| \sum_{i\in S}\sigma_i \| \theta_S^i \lambda_i({\bf J}^k_{:i} - \nabla f_i(x^*))\|^2,
$$
where $\sigma_i = \tfrac{1}{4L_i\beta_ip_i\lambda_i}$ for all $i$. If stepsize $\alpha$ satisfies 
\begin{equation}\label{alphaub1}
\alpha \leq \min_{1\leq i\leq n} \tfrac{p_i}{\mu + 4L_i\beta_i\lambda_ip_i},
\end{equation}
then 
$
\mathbb{E}[{\Psi}^k_S] \leq (1-\mu \alpha)^k \mathbb{E}[{\Psi}^0_S].
$
This implies that if we choose $\alpha$ equal to the upper bound in (\ref{alphaub1}), then   $\mathbb{E}[{\Psi}^k_S] \leq \epsilon \cdot \mathbb{E}[{\Psi}^0_S]$ as  long as 
$
k\geq \max_{i}\left\{ \tfrac{1}{p_i} + \tfrac{4L_i\beta_i\lambda_i }{\mu}\right\}  \log \left(\tfrac{1}{\epsilon}\right)  .
$
\end{theorem}

Our result involves a novel {\em stochastic} Lyapunov function ${\Psi}^k_S$, different from that  in \cite{Jacsketch}. We call this Lyapnuov function stochastic because it is a random variable when conditioned on $x^k$ and ${\bf J}_i^k$. Virtually all analyses of stochastic optimization algorithms in the literature (to the best of our knowledge, all except for that in \cite{Jacsketch}) involve deterministic Lyapunov functions. 

The result posits linear convergence with iteration complexity whose leading term involves the quantities $p_i$, $\beta_i$ (which depend on the sampling $S$ only)  and on $L_i$, $\mu$ and $\lambda_i$ (which depend on the properties and structure of $f$).

\subsection{Optimal bias-correcting random vector} Note that the complexity bound gets better as $\beta_i$ get smaller. Having said that, even for a fixed sampling $S$, the choice of $\beta_i$ is not unique, Indeed, this is because $\beta_i$ depends on the choice of $\theta_S$. In view of Lemma~\ref{lem:n98hf9hf}, we have many choices for this random vector.  Let $\Theta(S)$ be the collection of all bias-correcting random vectors associated with sampling $S$. In our next result we will compute the bias-correcting random vector $\theta_S$ which leads to the minimal complexity parameters $\beta_i$. In the rest of the paper, let
$
\mathbb{E}^i[\cdot] \eqdef \mathbb{E}[\cdot\;|\;i\in S].
$

\begin{lemma}\label{betaipro}
Let $S$ be a proper sampling. Then\\
(i) 
$
\min_{\theta \in \Theta(S)} \beta_i = \tfrac{1}{\sum_{C:i\in C}p_C/|C|} =  \tfrac{1}{p_i \mathbb{E}^i [1/ |S|]}
$
for all $i$, and the minimum is obtained at $\theta\in \Theta(S)$ given by $\theta_C^i = \tfrac{1}{|C| \sum_{C: i \in C} p_C/|C|} = \tfrac{1}{p_i|C| \mathbb{E}^i [1/ |S|]}$ for all $C: i\in C$; \\
(ii) 
$
\tfrac{1}{\mathbb{E}^i [1/ |S|]} \leq \mathbb{E}^i[|S|],
$
for all $i$. 
\end{lemma}

\subsection{Importance Sampling for Minibatches}\label{sec:ip}

In this part we construct an importance sampling for minibatches. This is in general a daunting task, and only a handful of papers exist on this topic. In particular, \citet{csiba2016importance} and \citet{ACD2019} focused on coordinate descent methods, and \citet{Jacsketch} considered minibatch SAGA with importance sampling over subsets of $[n]$ forming a partition.

Let $\tau: = \mathbb{E}[|S|]$ be the expected minibatch size. By Lemma~\ref{betaipro} and Theorem \ref{Th:convp}, if we choose the optimal ${\theta}_S$ as 
$
{\theta}_S^{i} = \tfrac{1}{p_i |S| \mathbb{E}^i[1/ |S|]}
$
for all $i$, then  $\beta_i = \tfrac{1}{p_i} \tfrac{1}{\mathbb{E}^i [1/ |S|]}$, and the iteration complexity bound becomes 
\begin{equation}\label{itbound}
\max_i\left\{ \tfrac{1}{p_i} + \tfrac{4L_i\lambda_i}{p_i \mu  \mathbb{E}^i[1/|S|]}\right\}  \log \left(\tfrac{1}{\epsilon}\right).
\end{equation}
Since $\mathbb{E}^i[1/ |S|]$ is nonlinear, it is difficult to minimize (\ref{itbound}) generally. Instead, we choose ${\theta}_S^{i}\equiv 1/p_i$ for all $i$, in which case $\beta_i = \mathbb{E}^i[|S|]/p_i$, and the complexity bound becomes 
\begin{equation}\label{itupbound1}
\max_i\left\{ \tfrac{1}{p_i} + \tfrac{4L_i\lambda_i\mathbb{E}^i[|S|]}{ p_i \mu }\right\} \log \left(\tfrac{1}{\epsilon}\right).
\end{equation}

From now on we focus on so-called {\em independent samplings.} Alternative results for {\em partition samplings} can be found in Section~\ref{sec:importance-partition} of the Supplementary. In particular, let $S$ be formed as follows: for each  $i\in [n]$ we flip a biased coin, independently, with probability of success $p_i>0$. If we are successful, we include $i$ in $S$. This type of sampling was also considered for nonconvex ERM problems in \cite{nonconvex_arbitrary} and for accelerated coordinate descent  in \cite{FilipACD18}. 

First, we calculate $\mathbb{E}^i[|S|]$ for this sampling $S$. Let the sampling $S^{\{i \}}$ includes every $j \in [n]\setminus \{ i \}$ in it, independently, with probability $p_j$. Then  
\[
\textstyle \mathbb{E}^i[|S|] = 1 + \mathbb{E}[|S^{\{i \}}|] =1 + \sum_{j \neq i} p_j 
=  \tau + 1 - p_i. 
\]
Since $\tau \leq \mathbb{E}^i[|S|] < \tau+1$, the complexity bound (\ref{itupbound1}) has an upper bound as 
\begin{equation}\label{itupbound2}
\max_i\left\{  \tfrac{\mu + 4L_i\lambda_i (\tau + 1)}{\mu p_i }\right\}  \log \left(\tfrac{1}{\epsilon}\right).
\end{equation}
Next, we minimize the upper bound (\ref{itupbound2}) over $p_1, p_2, ..., p_n$ under the condition $\mathbb{E}[|S|] = \tau$. Let 
$
q_i = \tfrac{\mu + 4L_i\lambda_i(\tau + 1)}{\sum_{j=1}^n [\mu + 4L_j\lambda_j (\tau +1)]} \cdot \tau, 
$
for $1\leq i \leq n$ and  denote $T = \{ i\;|\;q_i > 1 \}$. If $T = \emptyset$, then it is easy to see that $p_i = q_i$ minimize (\ref{itupbound2}), and consequently, (\ref{itupbound2}) becomes 
$
\textstyle \left(\tfrac{n}{\tau} + \tfrac{\sum_{i\in [n]}4L_i\lambda_i(\tau +1)}{\mu\tau} \right) \log \left(\tfrac{1}{\epsilon}\right).
$
Otherwise, in order to minimize (\ref{itupbound2}), we can choose $p_i = 1$ for $i \in T$, and $q_i \leq p_i \leq 1$ for $i \notin T$ such that $\sum_{i=1}^n p_i = \tau$. By choosing this optimal probability, (\ref{itupbound2}) becomes 
$
\textstyle \max_{i \in T} \left\{ \tfrac{\mu + 4L_i\lambda_i(\tau +1)}{\mu} \right\}  \log \left(\tfrac{1}{\epsilon}\right).
$

Notice that from Theorem \ref{Th:convp}, for $\tau$-nice sampling, the iteration complexity is 
\begin{equation}\label{eq:taunice}
\textstyle \left( \tfrac{n}{\tau} + \tfrac{4n \max_{i}L_i\lambda_i}{\mu} \right) \log \left(\tfrac{1}{\epsilon}\right). 
\end{equation}
Hence, compared to $\tau$-nice sampling, the iteration complexity for importance independent sampling could be at most $(1+1/\tau)$ times worse, but could also be $n/(\tau+1)$ times better in some extreme cases.

\subsection{SAGA-AS vs Quartz} \label{sec:SAGA-vsQuartz}

In this section, we compare our results for SAGA-AS with  known complexity results for the primal-dual method Quartz of \citet{Quartz}. We do this because this was the first and (with the exception of the dfSDCA method of \citet{dfSDCA}) remains the only SGD-type method for solving  \eqref{primal} which was analyzed in the arbitrary sampling paradigm. Prior to this work we have conjectured that SAGA-AS would attain the same complexity as Quartz. As we shall show, this is indeed the case.

The problem studied in \cite{Quartz}  is 
\begin{equation}\label{primal-phi-quartz}
\textstyle \min_{x\in \mathbb{R}^d} \tfrac{1}{n}\sum_{i=1}^n\phi_i({\bf A}_i^\top x)+ \psi(x) ,
\end{equation}
where ${\bf A}_i \in{ \mathbb{R}}^{d\times m}$, $\phi_i: {\mathbb{R}}^m \to \mathbb{R}$ is $1/\gamma$-smooth and convex, $\psi: \mathbb{R}^d \to \mathbb{R}$ is a $\mu$-strongly convex function.  When $\psi$ is also smooth, problem~\eqref{primal-phi-quartz}
can be written in the form of problem (\ref{primal}) with $\lambda_i = 1/n$, and 
\begin{equation}\label{eq:quartzfi}
\textstyle f_i(x) = \phi_i({\bf A}_i^\top x) +  \psi(x),
\end{equation}

Quartz guarantees the duality gap to be less than $\epsilon$ in expectation  using at most 
\begin{equation}\label{eq:iterationnoquartz}
\textstyle {\cal O}\left\{\max_{i} \left( \tfrac{1}{p_i} + \tfrac{v_i}{p_i\mu\gamma n} \right) \log\left( \tfrac{1}{\epsilon}\right)\right\}
\end{equation}
iterations, where the parameters $v_1, ..., v_n$ are assumed to satisfy the following expected separable overapproximation (ESO) inequality, which needs to hold for all $h_i\in \R^m$:
\begin{equation}\label{eq:ESOfirst}
\textstyle \mathbb{E}_S\left[ \left\|\sum_{i\in S} {\bf A}_ih_i \right\|^2 \right] \leq \sum_{i=1}^n p_iv_i\|h_i\|^2.
\end{equation}
If in addition $\psi$ is  $L_{\psi}$-smooth, then 
 $f_i$ in (\ref{eq:quartzfi})  is smooth  with  
$
L_i \leq \tfrac{\lambda_{\max}({\bf A}_i^{\top}{\bf A}_i)}{\gamma} +  {L_{\psi}}.
$ Let us now consider several particular samplings:

{\bf Serial samplings.} $S$ is said to be {\em serial} if $|S|=1$ with probability 1. So, $p_i = \Prob(S=\{i\})$ for all $i$.  By Lemma 5 in \cite{Quartz}, $v_i = \lambda_{\max}({\bf A}_i^{\top}{\bf A}_i)$. Hence the iteration complexity bound (\ref{eq:iterationnoquartz}) becomes 
$$
\textstyle {\cal O}\left\{\max_{i} \left( \tfrac{1}{p_i} + \tfrac{\lambda_{\max}({\bf A}_i^{\top}{\bf A}_i)}{p_i\mu\gamma n} \right) \log\left( \tfrac{1}{\epsilon}\right)\right\}.
$$
By choosing ${\theta}_S^{i} = 1/p_i$ (this is both the default choice mentioned in Lemma~\ref{lem:n98hf9hf}  and the optimal choice in view of Lemma~\ref{betaipro}), $\beta_i = 1/p_i$ and our iteration complexity bound in Theorem~\ref{Th:convp} becomes 
$$
\textstyle \max_{i}\left\{ \left(1+\tfrac{4L_{\psi}}{n\mu}\right)\tfrac{1}{p_i} + \tfrac{4\lambda_{\max}({\bf A}_i^{\top}{\bf A}_i)}{p_i \mu\gamma n}  \right\} \log\left(\tfrac{1}{\epsilon}\right).
$$
We can see that as long as $L_\psi/\mu = {\cal O}(n)$, the two bounds are essentially the same.

 {\bf Parallel ($\tau$-nice) sampling.} $S$ is said to be $\tau$-nice if it selects from all subsets of $[n]$ of cardinality $\tau$, uniformly at random. By Lemma 6 in \cite{Quartz}, 
$$
\textstyle v_i = \lambda_{\max}\left( \sum_{j=1}^d\left(1+ \tfrac{(\omega_j-1)(\tau-1)}{n-1}\right){\bf A}_{ji}^{\top}{\bf A}_{ji} \right),
$$
where ${\bf A}_{ji}$ is the $j$-th row of ${\bf A}_i$, and for each $1\leq j\leq d$, $\omega_j$ is the number of nonzero blocks in the $j$-th row of matrix ${\bf A}$, i.e., 
$
\omega_j \eqdef |\{ i\in [n]: {\bf A}_{ji} \neq 0 \}|, \quad 1\leq j \leq d.
$
In the dense case, i.e., $\omega_j = n$, $v_i = \tau \lambda_{\max}({\bf A}_i^{\top}{\bf A}_i)$. Hence, (\ref{eq:iterationnoquartz}) becomes 
$$
{\cal O}\left\{\max_{i} \left( \tfrac{n}{\tau} + \tfrac{\lambda_{\max}({\bf A}_i^{\top}{\bf A}_i)}{\mu\gamma } \right) \log\left( \tfrac{1}{\epsilon}\right)\right\}.
$$
By choosing ${\theta}_S^{i} = 1/p_i =n/ \tau$, we get $\beta_i = n$, and our iteration complexity bound in Theorem~\ref{Th:convp} becomes 
$$
\max_{i} \left\{ \tfrac{n}{\tau} + 4\tfrac{L_{\psi}}{\mu} + \tfrac{4\lambda_{\max}({\bf A}_i^{\top}{\bf A}_i)}{\mu \gamma}  \right\} \log \left(\tfrac{1}{\epsilon}\right).
$$
We can see the bounds are also essentailly the same up to some constant factors. However, if $\omega_j\ll n$, then Quartz enjoys a tighter bound.

The parameters $v_1,\cdots, v_n$ used in~\cite{Quartz} allow one to exploit the sparsity of the data matrix ${\bf A}=({\bf A}_1,\cdots, {\bf A}_n)$ and achieve almost linear speedup when ${\bf A}$ is sparse or has favourable spectral properties. In the next section, we study further SAGA-AS in the case when the objective function is of the form~\eqref{primal-phi-quartz}, and obtain results which, like Quartz, are able to improve with data sparsity.

\section{Analysis in the Composite Case}

We now consider the general problem \eqref{primal} with $\psi\neq 0$.

\subsection{Assumptions}

In order to be able to take advantage of data sparsity, we assume that functions $f_i$ take the form
\begin{align}\label{a:fphi}
f_i(x) \equiv  \phi_i({\bf A}_i^\top x).\end{align}
Then clearly 
$
\nabla f_i(x)={\bf A}_i \nabla \phi_i({\bf A}_i^\top x)$. Thus if SAGA-AS  starts with
$
\bJ^0=\begin{pmatrix} {\bf A}_1\alpha^0_1 &  {\bf A}_2\alpha^0_2&\cdots & {\bf A}_n\alpha^0_n\end{pmatrix},
$
for some $\alpha^0_i \in \R^m$, $i\in [n]$, then we always have
$
\bJ^k=\begin{pmatrix} {\bf A}_1\alpha^k_1 &  {\bf A}_2\alpha^k_2&\cdots & {\bf A}_n\alpha^k_n\end{pmatrix},
$ for some $\alpha^k_i \in \R^m$, $i\in [n]$. We assume that the set of minimizers
$
\cX^*\eqdef \arg\min\{P(x):x\in \R^d\}, 
$
is nonempty, and let $P^* = P(x^*)$ for $x^*\in \cX^*$.
Further, denote
$
[x]^*=\arg\min\{\|x-y\|:y\in \cX^*\};
$
 the closest optimal solution from $x$. Further, for any $M>0$  define $\cX(M)$
to be the set of points with objective value bounded by $P^*+M$, i.e.,
$$
\cX(M):=\{x\in \dom(\psi): P(x)\leq P^*+M)\}.
$$
We make several further assumptions:
\begin{assumption}[Smoothness]\label{ass:phi}
	Each $\phi_i:\R^m \rightarrow \R$ is $1/\gamma$-smooth and convex, i.e.,
	$$
	0\leq \langle \nabla \phi_i(a)-\nabla \phi_i(b), a-b\rangle \leq \|a-b\|^2/\gamma,\enspace \forall a,b\in \R^m.
	$$ 
\end{assumption}

\begin{assumption}[Quadratic functional growth condition; see~\cite{Necoara-Nesterov-Glineur-2018-linear-without-strong-convexity}]\label{ass:rsc}
	For any $M>0$, there is $\mu>0$ such that for any  $x\in\cX(M)$
	\begin{equation}
	\textstyle P(x)-P^*\geq \tfrac{\mu}{2}\|x- [x]^*\|^2. \label{eq:gc} 
	\end{equation}
\end{assumption}
\begin{assumption}[Nullspace consistency]\label{ass:uniquegradient}
	For any $x^*,y^*\in\cX^*$ we have $
	{\bf A}_i^\top x^*={\bf A}_i^\top y^*,\enspace \forall i\in [n].
	$
\end{assumption}

%\begin{algorithm}[ht]
%	%\caption{Minibatch SAGA with Arbitrary Sampling}
%	\caption{SAGA with Arbitrary Sampling}
%	\label{minibatchSAGA-prox}
%	\begin{algorithmic}
%		\STATE {\bfseries Parameters:} Sampling $S$ and $\theta_S$ satisfying Assumption~\ref{ass:thetaS}, stepsize $\alpha > 0$
%		\STATE {\bfseries Initialization:} Choose $x^0 \in \mathbb{R}^d$, ${\bf J}^0 \in \mathbb{R}^{d\times n}$ 
%		\FOR{ $k = 0, 1, 2, ...$}
%		\STATE Sample a fresh set $S_k \sim S$ 
%		\STATE ${\bf J}^{k+1} = {\bf J}^k + ({\bf G}(x^k) - {\bf J}^k){\bf \Pi}_{{\bf I}_{S_k}}$ 
%		\STATE $g^k = {\bf J}^k\lambda + ({\bf G}(x^k) - {\bf J}^k){\theta_{S_k}}{\bf \Pi}_{{\bf I}_{S_k}} \lambda$ 
%		\STATE $x^{k+1} =\prox^\psi_{\alpha}\left( x^k -\alpha g^k\right)$ 
%		\ENDFOR
%	\end{algorithmic}
%\end{algorithm}

\subsection{Linear convergence under quadratic functional growth condition}

%The following almost sure convergence result, which was proved for SAGA, still holds in the mini batch setting.
%\begin{theorem}\label{th:poon18}[\citet{pmlr-v80-poon18a}.]
%If $\alpha\leq \tfrac{1}{3n\max_i {\lambda_i L_i}}$,
%The sequence of $\{P(x^k)\}$ almost surely converges to $P^*$.
%\end{theorem}
%\zheng{I will check later  the correct stepsize setting.}
%\peter{Do we want this result?}

Our first complexity result states a linear convergence rate of SAGA-AS under the quadratic functional growth condition.

\begin{theorem}\label{th:phi}
	Assume $f$ is $L$-smooth. Let $v\in \R^n_+$ be a positive vector satisfying~\eqref{eq:ESOfirst} for a proper sampling $S$.
%	\begin{equation} \label{eq:ESOfirst}  \textstyle
%	\Exp_{\hat{S}} \left[ \left\| \sum_{i\in \hat{S} } {\bf A}_i h_i \right\|^2 \right] \leq   \sum_{i=1}^n p_i v_i \|h_i\|^2.
%	\end{equation} 
	Let $\{ x^k, {\bf J}^k \}$ be the iterates produced by Algorithm \ref{minibatchSAGA} with ${\theta}_S^{i} = 1/p_i$ for all $i$ and $S$. Let any $x^*\in \cX^*$. Consider the  Lyapunov function 
	\begin{align*}
	\Psi^k  &\eqdef \|x^k -[x^k]^*\|^2 
	\\& \textstyle \qquad + \alpha \sum_{i=1}^n\sigma_i  p_i^{-1}v_i\lambda_i^2 \|\alpha_i^{k}-\nabla \phi_i({\bf A}_i^\top x^*)\|^2,
	\end{align*}
	where $\sigma_i = \tfrac{\gamma}{2v_i\lambda_i}$ for all $i$. Then there is a constant $\mu>0$ such that the following is true. If stepsize $\alpha$ satisfies 
	\begin{equation}\label{alphaub}
	\alpha\leq \min\left\{\tfrac{2}{3}\min_{1\leq i\leq n} \tfrac{p_i}{\mu + 4 v_i\lambda_i/\gamma}, \tfrac{1}{3L} \right\},
	\end{equation}
	then 
	$
	\mathbb{E}[{\Psi}^k] \leq \left( \tfrac{1+\alpha\mu/2}{1+\alpha\mu}\right)^k \mathbb{E}[{\Psi}^0].
	$
	This implies that if we choose $\alpha$ equal to the upper bound in (\ref{alphaub}), then $\mathbb{E}[{\Psi}^k] \leq \epsilon \cdot \mathbb{E}[{\Psi}^0]$ when
	$$
	k \geq \left(2 + \max\left\{ \tfrac{6L}{\mu}, 3\max_{i} \left( \tfrac{1}{p_i} + \tfrac{4v_i\lambda_i}{ p_i \mu \gamma} \right)  \right\} \right) \log \left(\tfrac{1}{\epsilon}\right).
	$$
\end{theorem}

Also note that the Lyapunov function considered is {\em not stochastic} (i.e., $\Exp[\Psi^k]\;|\; x^k, \alpha_i^k]$ is not random).

Non-strongly convex problems in the form of~\eqref{primal} and~\eqref{a:fphi} which satisfy Assumptions~\ref{ass:phi}, ~\ref{ass:rsc} and~\ref{ass:uniquegradient}  include the case when each $\phi_i$ is strongly convex and  $\psi$ is polyhedral~\cite{Necoara-Nesterov-Glineur-2018-linear-without-strong-convexity}. In particular, Theorem~\ref{th:phi}~applies to the following logistic regression problem that we use in the experiments ($\lambda_1 \geq 0$ and $\lambda_2\geq 0$)
\begin{align}\label{a:log}
 \min_{x\in \R^d} \sum_{i=1}^n \log(1+e^{b_iA_i^\top x})+\lambda_1 \|x\|_1+\tfrac{\lambda_2}{2} \|x\|^2.
\end{align}
%for any value of $\lambda_1\geq 0$ and $\lambda_2\geq 0$.

\begin{remark}\label{rem:saga}It is important to note that  the estimation of the  constant $\mu$ in Theorem~\ref{th:phi} is not an easy task.  However we can obtain $\alpha$ satisfying~\eqref{alphaub} without any knowledge of $\mu$ and keep the same  order of complexity bound. W.l.o.g. we can assume $\mu\leq 4v_i\lambda_i/\gamma$. Then~\eqref{alphaub} is satisfied by taking $
\alpha= \min\left\{\tfrac{1}{12}\min_{1\leq i\leq n} \tfrac{p_i \gamma}{v_i\lambda_i}, \tfrac{1}{3L} \right\}
$. This makes SAGA stand out from all the stochastic variance reduced methods including SDCA~\cite{SDCA}, SVRG~\cite{prox-SVRG}, SPDC~\cite{SPDC}, etc. We also note that an adaptive strategy was proposed for SVRG in order to avoid guessing the constant $\mu$~\cite{AdapSVRG}. Their adaptive approach leads to a triple-looped algorithm (as SVRG is already double-looped) and  linear convergence  as we have in Theorem~\ref{th:phi} was not obtained.
\end{remark}

\subsection{Linear convergence for strongly convex regularizer}

For the problem studied in \cite{Quartz} where the regularizer $\psi$ is $\mu$-strongly convex (and hence Assumption~\eqref{ass:rsc} holds and the minimizer is unique), we  obtain the following refinement of Theorem~\ref{th:phi}.

\begin{theorem}\label{th:phistrongly}
	Let $\psi$ be $\mu$-strongly convex. Let $v\in \R^n_+$ be a positive vector satisfying~\eqref{eq:ESOfirst} for a proper sampling $S$.
	Let $\{ x^k, {\bf J}^k \}$ be the iterates produced by Algorithm \ref{minibatchSAGA} with ${\theta}_S^{i} = 1/p_i$ for all $i$ and $S$. Consider the same Lyapunov function as in Theorem~\ref{th:phi}, but set
 $\sigma_i = \tfrac{2\gamma}{3v_i\lambda_i}$ for all $i$. If stepsize $\alpha$ satisfies 
	\begin{equation}\label{alphaubstrongly}
	\textstyle \alpha \leq \min_{1\leq i\leq n} \tfrac{p_i}{\mu + 3v_i\lambda_i/\gamma},
	\end{equation}
	then 
	$
	 \mathbb{E}[{\Psi}^k] \leq \left( \tfrac{1}{1+\alpha\mu}\right)^k \mathbb{E}[{\Psi}^0].
	$
	This implies that if we choose $\alpha$ equal to the upper bound in (\ref{alphaubstrongly}), then $\mathbb{E}[{\Psi}^k] \leq \epsilon \cdot \mathbb{E}[{\Psi}^0]$ when
	$
	\textstyle k \geq \max_{i} \left\{ 1+ \tfrac{1}{p_i} + \tfrac{3v_i\lambda_i}{p_i\mu \gamma} \right\} \log \left(\tfrac{1}{\epsilon}\right).
	$
\end{theorem}

Note that up to some small constants, the rate provided by Theorem~\ref{th:phistrongly} is the same as that of Quartz. Hence, the analysis for special samplings provided in Section~\ref{sec:SAGA-vsQuartz} applies, and we conclude that {\em SAGA-AS is also able to accelerate on sparse data.}

\section{Experiments}

%\peter{We need to mention all parameters we choose for the results to be reproducible. Can do it in Supplementary if there is no space here; but this has ti be clear.}

We tested SAGA-AS to solve the logistic regression problem~\eqref{a:log} on three different datasets:   w8a, a9a and ijcnn1\footnote{https://www.csie.ntu.edu.tw/~cjlin/libsvmtools/datasets/}. The experiments presented in Section~\ref{sec:BS} and~\ref{sec:IP} are tested for $\lambda_1=0$ and $\lambda_2=1e-5$, which is of the same order as the number of samples in the three datasets.  In Section~\ref{sec:CD} we test on the unregularized problem with $\lambda_1=\lambda_2=0$. More experiments can be found in the Supplementary.

Note that in all the plots, the x-axis records the number of pass of the dataset, which by theory should grow no faster than $O(\tau /(\alpha\mu)\log(1/\epsilon))$ to obtain $\epsilon$-accuracy.

\subsection{Batch sampling}\label{sec:BS}
\begin{figure}[!h]
  \centering 
  \subfigure[SAGA; ijcnn1]{ 
    \label{fig:subfig:2a} %% label for first subfigure 
    \includegraphics[height=0.2\textwidth]{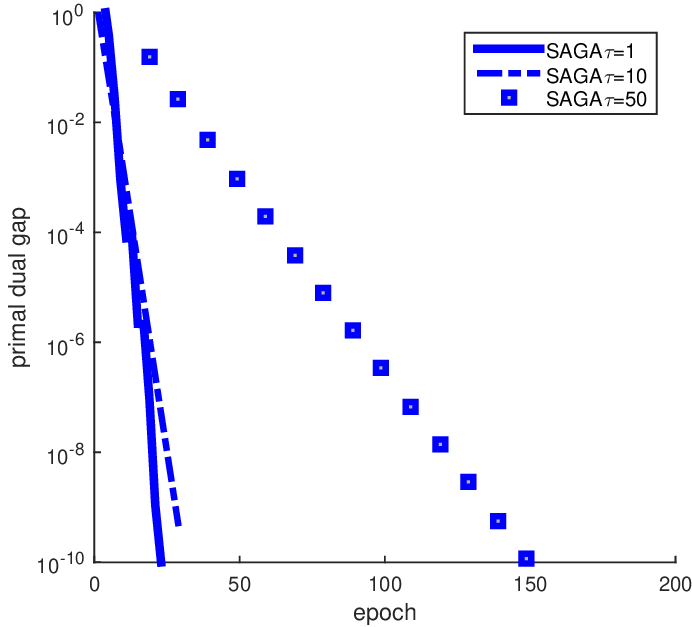}} 
  \subfigure[SDCA; ijcnn1]{ 
    \label{fig:subfig:2b} %% label for second subfigure 
    \includegraphics[height=0.2\textwidth]{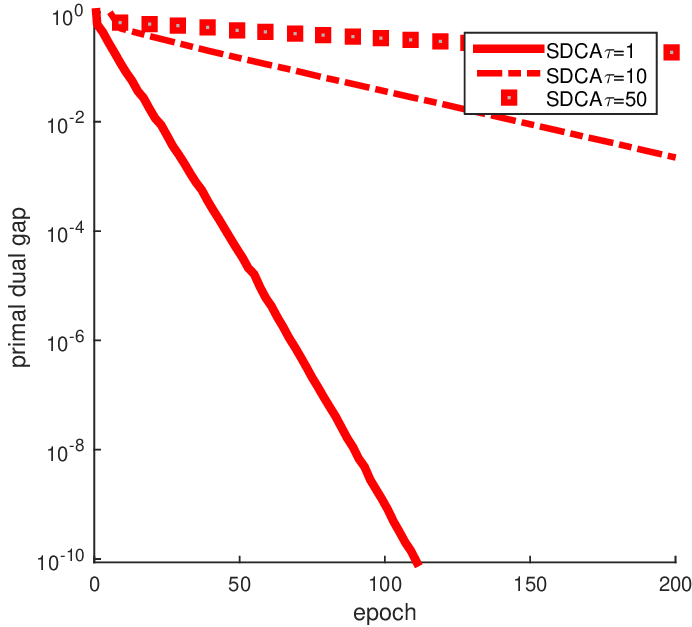}} 
  \caption{ mini-batch SAGA V.S. mini-batch SDCA}
  \label{fig1} %% label for entire figure 
\end{figure}

Here we compare SAGA-AS with  SDCA in the case when $S$ is a $\tau$-nice sampling and for $\tau\in \{1,10,50\}$. Note that SDCA with $\tau$-nice sampling works  the same both in theory and in practice as Quartz with $\tau$-nice sampling.  We report in Figure~\ref{fig1} the results obtained for the dataset ijcnn1.  When we increase $\tau$ by 50, the number of epochs of SAGA-AS  only increased by less than 6. This indicates a considerable speedup if parallel computation is used in the implementation of mini-batch case.

\subsection{Importance sampling}\label{sec:IP}

\begin{figure} [!h]
  \centering 
  \subfigure[ijcnn1]{ 
    \label{fig:subfig:1a} %% label for first subfigure 
    \includegraphics[height=0.2\textwidth]{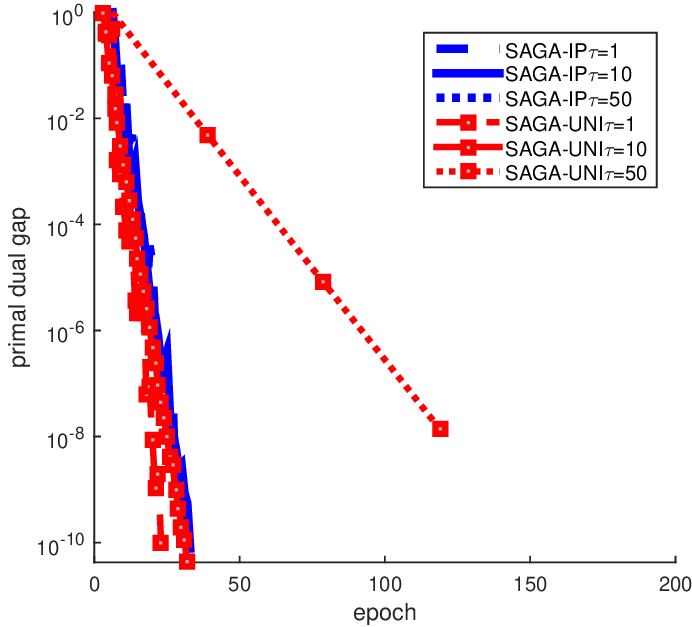}} 
  \subfigure[w8a]{ 
    \label{fig:subfig:1b} %% label for second subfigure 
    \includegraphics[height=0.2\textwidth]{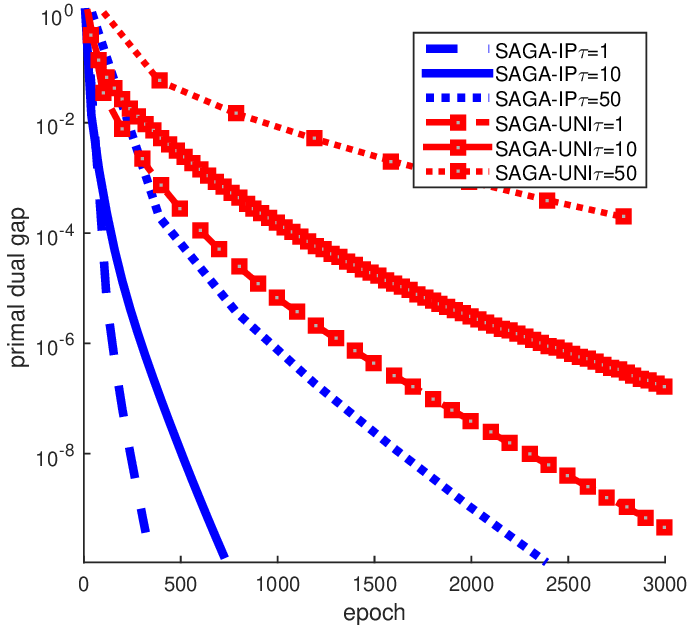}} 
  \caption{importance sampling V.S. uniform sampling} 
  \label{fig2} %% label for entire figure 
\end{figure}

We compare uniform sampling SAGA (SAGA-UNI) with importance sampling SAGA (SAGA-IP), as described in Section~\ref{sec:ip}
, on three values of $\tau\in\{1,10, 50\}$.   The results for the datasets w8a and ijcnn1 are shown in Figure~\ref{fig2}. 
For the dataset ijcnn1, mini-batch with importance sampling almost achieves linear speedup as the number of epochs does not increase with $\tau$. For the dataset w8a, mini-batch with importance sampling can even need less number of epochs than serial uniform sampling. Note that we adopt the importance sampling strategy described in~\cite{ACD2019} and the actual running time is the same as uniform sampling.

\subsection{Comparison with coordinate descent}\label{sec:CD}

We consider the un-regularized logistic regression problem~\eqref{a:log} with $\lambda_1=\lambda_2=0$. In this case, Theorem~\ref{th:phi} applies and we expect to have linear convergence of SAGA without any knowledge on the constant $\mu$ satsifying Assumption~\eqref{ass:rsc}, see Remark~\ref{rem:saga}. This makes SAGA comparable with descent methods such as gradient method and coordinate descent (CD) method. However, comparing with their deterministic counterparts, the speedup provided by CD can be at most of order $d$ while  the speedup by SAGA can be of order $n$.  Thus SAGA is much preferable  than CD when $n$ is larger than $d$.  We provide numerical evidence in Figure~\ref{fig5}.

\begin{figure} [!h]
  \centering 
  \subfigure[ijcnn1]{ 
    \label{fig:subfig:5a} %% label for first subfigure 
    \includegraphics[height=0.2\textwidth]{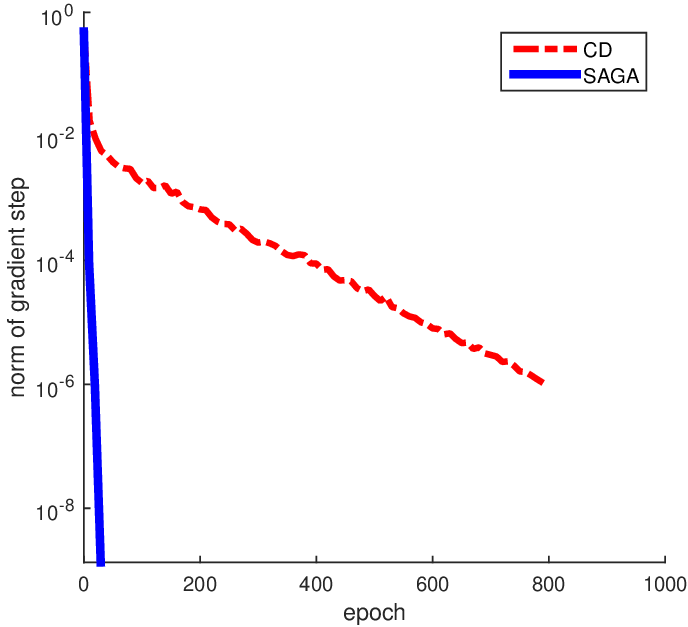}} 
     \subfigure[a9a]{ 
    \label{fig:subfig:5c} %% label for second subfigure 
    \includegraphics[height=0.2\textwidth]{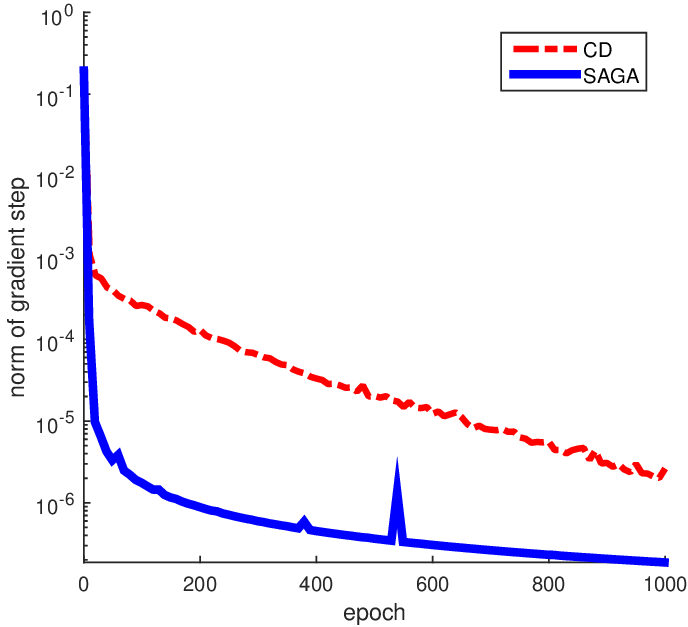}} 
  \caption{SAGA V.S. CD} 
  \label{fig5} %% label for entire figure 
\end{figure}

\iffalse
\begin{figure} [!h]
  \centering 
  \subfigure[ijcnn1]{ 
    \label{fig:subfig:5a} %% label for first subfigure 
    \includegraphics[height=0.2\textwidth]{plots2/ijcnn1vsCD.eps}} 
  \subfigure[w8a]{ 
    \label{fig:subfig:5b} %% label for second subfigure 
    \includegraphics[height=0.2\textwidth]{plots2/w8avsCD.eps}} 
     \subfigure[a9a]{ 
    \label{fig:subfig:5c} %% label for second subfigure 
    \includegraphics[height=0.2\textwidth]{plots2/a9avsCD.eps}} 
  \caption{ SAGA with CD} 
  \label{fig5} %% label for entire figure 
\end{figure}

\fi


%\section*{Acknowledgements}
%
%\textbf{Do not} include acknowledgements in the initial version of
%the paper submitted for blind review.
%
%If a paper is accepted, the final camera-ready version can (and
%probably should) include acknowledgements. In this case, please
%place such acknowledgements in an unnumbered section at the
%end of the paper. Typically, this will include thanks to reviewers
%who gave useful comments, to colleagues who contributed to the ideas,
%and to funding agencies and corporate sponsors that provided financial
%support.

\nocite{langley00}

\bibliography{SAGA_Abitrary_Sampling}
%\bibliography{example_paper}
\bibliographystyle{icml2019}


%%%%%%%%%%%%%%%%%%%%%%%%%%%%%%%%%%%%%%%%%%%%%%%%%%%%%%%%%%%%%%%%%%%%%%%%%%%%%%%
%%%%%%%%%%%%%%%%%%%%%%%%%%%%%%%%%%%%%%%%%%%%%%%%%%%%%%%%%%%%%%%%%%%%%%%%%%%%%%%

\appendix
\onecolumn

\icmltitle{SUPPLEMENTARY MATERIAL \\ SAGA with Arbitrary Sampling}

\section{Importance Sampling for Minibatches: Partition Sampling} \label{sec:importance-partition}

In this section, we consider SAGA-AS with partition sampling for the problem 
\begin{equation}\label{problemn}
x^* = \arg \min_{x\in \mathbb{R}^d} \left[ f(x) \eqdef \sum_{i=1}^n \frac{1}{n} f_i(x)\right].
\end{equation}
First we give the definition of partition of $[n]$ and partition sampling. 

\begin{definition}
A partition ${\cal G}$ of $[n]$ is a set consisted of the subsets of $[n]$ such that $\cup_{C\in {\cal G}} C = [n]$ and $C_i \cap C_j = \emptyset$ for any $C_i$, $C_j\in {\cal G}$ with $i\neq j$. A partition sampling $S$ is a sampling such that $p_C = \mathbb{P}[S=C] > 0$ for all $C \in {\cal G}$ and $\sum_{C \in {\cal G}} p_C = 1$. 
\end{definition}

From (\ref{eq:nbdfg98ud8df}) of Lemma \ref{lem:n98hf9hf}, for partition sampling, we have ${\theta}_C^i = \frac{1}{p_C}$ if $i\in C$ for all $C \in {\cal G}$. Hence, SAGA-AS for problem (\ref{problemn}) with partition sampling becomes Algorithm \ref{minibatchSAGAps} (SAGA-PS):

\begin{algorithm}[h]
	\caption{SAGA with Partition Sampling for problem (\ref{problemn})--SAGA-PS}
	\label{minibatchSAGAps}
	\begin{algorithmic}
		\STATE {\bfseries Parameters:} Partition sampling $S$ and stepsize $\alpha > 0$
		\STATE {\bfseries Initialization:} Choose $x^0 \in \mathbb{R}^d$, ${\bf J}^0 \in \mathbb{R}^{d\times n}$ 
		\FOR{ $k = 0, 1, 2, ...$}
		\STATE Sample a fresh set $S_k \sim S$ 
		\STATE ${\bf J}^{k+1} = {\bf J}^k + \left({\bf G}(x^k) - {\bf J}^k\right){\bf \Pi}_{{\bf I}_{S_k}}$ 
		\STATE $g^k = {\bf J}^k\lambda + \frac{1}{np_{S_k}}\left({\bf G}(x^k) - {\bf J}^k\right){\bf \Pi}_{{\bf I}_{S_k}}e$ 
		\STATE $x^{k+1} = x^k -\alpha g^k$ 
		\ENDFOR
	\end{algorithmic}
\end{algorithm}

Next we will give the iteration complexity of SAGA-PS by reformulation. For any partition sampling $S$, let $f_C(x) = \frac{1}{|C|}\sum_{i\in C} f_i(x)$ for $C\in {\cal G}$, and let $f_C$ be $L_C$-smooth. In problem (\ref{problemn}), 
$
\textstyle f(x) = \frac{1}{n}\sum_{i=1}^n f_i(x) = \frac{1}{n}\sum_{C \in {\cal G}} |C| f_C(x) =  \sum_{C\in {\cal G}} \frac{|C|}{n} f_C(x).
$
 Let $m = |{\cal G}|$ and without loss of generality, we denote ${\cal G} = \{ C_1, C_2, ..., C_m \}$. Let $\lambda_i = \frac{|C_i|}{n}$ for $1\leq i \leq m$. Then we can see the minibatch SAGA with partition sampling for problem (\ref{problemn}) (Algorithm \ref{minibatchSAGAps}) can be regarded as Algorithm \ref{minibatchSAGA} for problem (\ref{primal}) with the sampling: $\mathbb{P}[{\{i \}}] = p_{C_i}$ and ${\theta}_{\{i \}}^j= \frac{1}{p_{C_i}}$ for all $i$, $j\in [m]$. Hence, by applying Theorem \ref{Th:convp}, we can obtain the following theorem. 
 
 \begin{theorem}\label{Th:convps}
 	Let $S$ be any partition sampling with partition ${\cal G} = \{ C_1, C_2, ..., C_m \}$. Let $f$ in problem (\ref{problemn}) be $\mu$-strongly convex and $f_C$ be convex and $L_C$-smooth. Let $\{x^k, {\bf J}^k\}$ be the iterates produced by Algorithm \ref{minibatchSAGAps}. \\
 	Consider the stochastic Lyapunov function 
 	$$
 	\Psi^k_{S}\eqdef \|x^k - x^*\|^2 + 2\alpha \sigma_{S} \left\|\frac{1}{np_{S}} ({\bf J}^k - {\bf G}(x^k)){\Pi}_{{\bf I}_{S}}e \right\|^2,
 	$$
 	where $\sigma_{S} = \frac{n}{4L_{S}|S|}$ is a stochastic Lyapunov constant. If stepsize $\alpha$ satisfies 
 		\begin{equation}\label{alphaps}
 	\alpha \leq \min_{C\in {\cal G}} \frac{p_C}{\mu + \frac{4L_C|C|}{n}},
 	\end{equation}
 	then 
 	$
 	\mathbb{E}[{\Psi}^k_S] \leq (1-\mu \alpha)^k \mathbb{E}[{\Psi}^0_S].
 	$
 	This implies that if we choose $\alpha$ equal to the upper bound in (\ref{alphaps}), then 
 	$$
 	k\geq \max_{C\in {\cal G}}\left\{ \frac{1}{p_C} + \frac{4L_C|C|}{\mu np_C} \right\}  \log \left(\frac{1}{\epsilon}\right) \Rightarrow  \mathbb{E}[{\Psi}^k_S] \leq \epsilon \cdot \mathbb{E}[{\Psi}^0_S].
 	$$ 
 \end{theorem}
Theorem \ref{Th:convps} contains Theorem 5.2 with $\tau$-partition sampling  in \cite{Jacsketch} as a special case, and with a little weaker condition: instead of demanding $f_C$ be $\mu$-strongly convex, we only need $f$ be $\mu$-strongly convex. 

\subsection{Partition sampling}

From Theorem \ref{Th:convps}, we can propose importance partition sampling for Algorithm \ref{minibatchSAGAps}. For a partition sampling $S$ with the partition ${\cal G} = \{C_1, C_2, ..., C_m \}$, where $m = |{\cal G}|$, the iteration complexity of Algorithm~\ref{minibatchSAGAps} is given by 
$$
\max_{C\in {\cal G}}\left\{ \frac{1}{p_C} + \frac{4L_C|C|}{\mu np_C} \right\} \log \left(\frac{1}{\epsilon}\right).
$$
We can minimize the complexity bound in $p_{C}$ by choosing 
\begin{equation}
p_{C} = \frac{\mu n + 4L_C|C|}{\sum_{C\in {\cal G}}(\mu n + 4L_C|C|)}.
\end{equation}
With these optimal probabilities, the stepsize bound is $\alpha \leq \frac{n}{\sum_{C\in {\cal G}}(\mu n + 4L_C|C|)}$, and by choosing the maximum allowed stepsize the resulting complexity becomes 
\begin{equation}
\left( |{\cal G}| + \frac{4\sum_{C\in {\cal G}}L_C|C|}{\mu n} \right) \log \left(\frac{1}{\epsilon}\right).
\end{equation}

\newpage
\section{Proofs of Lemmas ~\ref{lem:n98hf9hf} and ~\ref{betaipro} }

\subsection{Proof of Lemma~\ref{lem:n98hf9hf}}

From (\ref{eq:nb98fg98bf}), we know (\ref{eq:BC-cond}) in Assumption \ref{ass:thetaS} is actually  
$$
\mathbb{E}[\theta_S {\rm Diag}(e_S)\lambda] = \lambda, 
$$
which is equivalent to 
$$
\mathbb{E}[{\theta}^i_S 1_{i\in S}\lambda_i] = \lambda_i,
$$
for all $i\in [n]$. Then for all $i\in [n]$, we have 
\begin{eqnarray*}
\mathbb{E}[{\theta}^i_S 1_{i\in S}\lambda_i] &=& \sum_{C\subseteq [n]}p_C {\theta}^i_C 1_{i\in C} \lambda_i \\ 
&=& \sum_{C\subseteq [n]: i\in C} p_C {\theta}^i_C \lambda_i \\ 
&=& \lambda_i.
\end{eqnarray*}
Since $\lambda_i >0$ for all $i\in [n]$, we have that (\ref{eq:BC-cond}) is equivalent to 
$$
\sum_{C\subseteq [n]: i\in C} p_C {\theta}^i_C = 1, \quad \forall i\in [n].
$$

\subsection{Proof of Lemma~\ref{betaipro}(i)}

From Lemma \ref{lem:n98hf9hf}, the problem $\min_{\theta \in {\Theta}(S)} \beta_i$ is equivalent to the following linearly constrained convex problem: 
\begin{equation} \label{betaisubp}
\begin{array}{rcl}
\min & & \beta_i = \sum_{C: i\in C}p_C|C|({\theta}_C^i)^2 \\
{\rm s.t.\ }& & \sum_{C: i\in C}p_C {\theta}_C^i = 1.
\end{array}
\end{equation}
The KKT system of problem (\ref{betaisubp}) is 
\begin{equation}\label{kktbetai}
\left\{
\begin{array}{l}
2p_C|C|{\theta}_C^i + {\tilde \lambda} p_C = 0, \forall C: i\in C\\
\sum_{C: i\in C} p_C {\theta}_C^i = 1,
\end{array}\right.
\end{equation}
where ${\tilde \lambda} \in \mathbb{R}$ is the Lagrangian dual variable. By solving system (\ref{kktbetai}), we obtain the optimal solution 
$$
{\theta}_C^i = \frac{1}{|C| \sum_{C: i \in C}\frac{p_C}{|C|}} = \frac{1}{p_i|C| \mathbb{E}^i[\frac{1}{|S|}]},
$$
for all $C: i\in C$,  and the minimum of $\beta_i$ is 
\begin{eqnarray*}
\beta_i &=& \sum_{C: i\in C} p_C |C| ({\theta}_C^i)^2 \\
&=& \sum_{C: i\in C} p_C|C|\frac{1}{(p_i|C| \mathbb{E}^i[\frac{1}{|S|}])^2} \\
&=& \frac{1}{p_i}\cdot \frac{1}{(\mathbb{E}^i[\frac{1}{|S|}])^2} \sum_{C: i\in C} \frac{p_C}{p_i}\frac{1}{|C|} \\
&=& \frac{1}{p_i}\cdot \frac{1}{\mathbb{E}^i[\frac{1}{|S|}]}.
\end{eqnarray*}

\subsection{Proof of Lemma~\ref{betaipro}(ii)}

From (\ref{eq:nbdfg98ud8df}) of Lemma \ref{lem:n98hf9hf}, we can choose ${\theta}_C^i = \frac{1}{p_i}$ for all $i$ and $C$. Hence 
$$
\min_{\theta \in {\Theta}(S)} \beta_i \leq \sum_{C: i\in C}p_C |C| \frac{1}{p_i^2} = \frac{1}{p_i}\mathbb{E}^i[|S|],
$$
which implies 
$$
\frac{1}{\mathbb{E}^i[\frac{1}{|S|}]} \leq \mathbb{E}^i[|S|].
$$

\newpage
\section{Smooth Case: Proof of Theorem~\ref{Th:convp}}

\subsection{Lemmas}

The following inequality is a direct consequence of convexity  of $x\mapsto \|x\|^2$.
\begin{lemma}\label{lm:ineq}
Let $a^i \in \mathbb{R}^d$ for $1\leq i \leq z$ with $z \geq 1$. Then 
$$
\left\| \sum_{i=1}^z a^i \right\|^2 \leq z \sum_{i=1}^z \|a^i\|^2.
$$	
\end{lemma}

\begin{lemma}\label{lm:Jk+1}
Let $S$ be a sampling, and $\sigma_i$ be any non-negative constant for $1\leq i\leq n$. Then 
\begin{eqnarray*}
\mathbb{E}\left[ |S|\sum_{i\in S}\sigma_i \left\|{\theta}_S^i\lambda_i ({\bf J}^{k+1}_{:i} - \nabla f_i(x^*))\right\|^2 \right] 
& \leq & \mathbb{E}\left[ |S|\sum_{i\in S}(1-p_i)\sigma_i \left\|{\theta}_S^i\lambda_i ({\bf J}^{k}_{:i} - \nabla f_i(x^*))\right\|^2 \right] \\
&& \quad + \mathbb{E}\left[ \sum_{i=1}^n \sigma_ip_i\beta_i\lambda_i^2 \left\|\nabla f_i(x^k) - \nabla f_i(x^*) \right\|^2 \right].
\end{eqnarray*}
\end{lemma}
\begin{proof}
First notice that 
$$
{\bf J}^{k+1}_{:i} = \left\{ \begin{array}{rl}
\nabla f_i(x^k), & \mbox{ if $i \in { S^k}$} \\
{\bf J}^k_{:i}. &\mbox{ if $i \notin S^k$ }
\end{array} \right.
$$
Then by taking conditional expectation on ${\bf J}^k$ and $x^k$, we have 
\begin{eqnarray*}
&&\mathbb{E} \left[ |S| \sum_{i\in S} \sigma_i \left\|{\theta}_S^i \lambda_i ({\bf J}^{k+1}_{:i} - \nabla f_i(x^*)) \right\|^2\;|\;{\bf J}^k, x^k \right] \\
&=& \sum_{C}\sum_{\tilde C} \sum_{i \in C} p_C p_{\tilde C} |C|\sigma_i \left\|{\theta}_C^i \lambda_i ({\bf J}^{k+1}_{:i}|_{S^k = \tilde C} - \nabla f_i(x^*)) \right\|^2 \\
&=& \sum_{C}p_C\sum_{i\in C}|C| \sigma_i ({\theta}_C^i)^2 \lambda_i^2 \sum_{\tilde C}p_{\tilde C}\left\|{\bf J}^{k+1}_{:i}|_{S^k = \tilde C} - \nabla f_i(x^*)\right\|^2 \\
&=& \sum_{C}p_C \sum_{i\in C} |C| \sigma_i ({\theta}_C^i)^2 \lambda_i^2 \left[ \sum_{{\tilde C} : i \in {\tilde C}} p_{\tilde C} \left\| \nabla f_i(x^k) - \nabla f(x^*)\right\|^2 + \sum_{{\tilde C}: i\notin {\tilde C}} p_{\tilde C} \left\|{\bf J}^k_{:i} - \nabla f_i(x^*)\right\|^2 \right] \\
&=& \sum_{C}p_C \sum_{i\in C} |C| \sigma_i ({\theta}_C^i)^2 \lambda_i^2 \left[ p_i \left\| \nabla f_i(x^k) - \nabla f(x^*)\right\|^2 + (1-p_i) \left\|{\bf J}^k_{:i} - \nabla f_i(x^*)\right\|^2 \right] \\
&=& \mathbb{E} \left[ |S| \sum_{i\in S} (1-p_i)\sigma_i \left\|{\theta}_S^i \lambda_i ({\bf J}^{k}_{:i} - \nabla f_i(x^*)) \right\|^2\;|\;{\bf J}^k, x^k \right] + \sum_C p_C \sum_{i\in C} |C| \sigma_i p_i ({\theta}_C^i)^2 \lambda_i^2 \left\|\nabla f_i(x^k) - \nabla f_i(x^*) \right\|^2 \\
&=& \mathbb{E} \left[ |S| \sum_{i\in S} (1-p_i)\sigma_i \left\|{\theta}_S^i \lambda_i ({\bf J}^{k}_{:i} - \nabla f_i(x^*)) \right\|^2\;|\;{\bf J}^k, x^k \right] + \sum_{i=1}^n \sigma_i p_i\lambda_i^2 \sum_{C: i\in C} p_C |C| ({\theta}_C^i)^2 \left\|\nabla f_i(x^k) - \nabla f(x^*) \right\|^2 \\
&=& \mathbb{E} \left[ |S| \sum_{i\in S} (1-p_i)\sigma_i \left\|{\theta}_S^i \lambda_i ({\bf J}^{k}_{:i} - \nabla f_i(x^*)) \right\|^2\;|\;{\bf J}^k, x^k \right] + \sum_{i=1}^n \sigma_i p_i \beta_i \lambda_i^2 \left\| \nabla f_i(x^k) - \nabla f_i(x^*) \right\|^2 .
\end {eqnarray*}

Taking expectations again and applying the tower property, we obtain the result. 

\end{proof}

In the next lemma we bound the second moment of the gradient estimate $g^k$. 

\begin{lemma}\label{lm:gk2}
The second moment of the gradient estimate is bounded by 
\begin{eqnarray*}
\mathbb{E}\left[ \|g^k\|^2\;|\;{\bf J}^k, x^k \right] 
& \leq & 2\mathbb{E}\left[ |S|\sum_{i\in S} \left\|{\theta}_S^i\lambda_i ({\bf J}^{k}_{:i} - \nabla f_i(x^*))\right\|^2 \right]  + 2\sum_{i=1}^n \beta_i\lambda_i^2 \left\|\nabla f_i(x^k) - \nabla f_i(x^*) \right\|^2.
\end{eqnarray*}
\end{lemma}

\begin{proof}
Recall that 
\begin{eqnarray*}
g^k &=& {\bf J}^k \lambda + ({\bf G}(x^k) - {\bf J}^k){\theta}_{S_k}{\bf \Pi}_{{\bf I}_{S_k}} \lambda \\
&=& {\bf J}^k \lambda - ({\bf J}^k - {\bf G}(x^*)){\theta}_{S_k}{\bf \Pi}_{{\bf I}_{S_k}}\lambda + ({\bf G}(x^k) - {\bf G}(x^*)){\theta}_{S_k}{\bf \Pi}_{{\bf I}_{S_k}}\lambda.
\end{eqnarray*}
By applying Lemma \ref{lm:ineq} with $z=2$, we get 
$$
\left\|g^k\right\|^2 \leq 2 \left\| ({\bf G}(x^k) - {\bf G}(x^*)){\theta}_{S_k}{\bf \Pi}_{{\bf I}_{S_k}}\lambda \right\|^2 + 2 \left\| ({\bf J}^k - {\bf G}(x^*)){\theta}_{S_k}{\bf \Pi}_{{\bf I}_{S_k}}\lambda - {\bf J}^k \lambda \right\|^2, 
$$
which implies that 
\begin{eqnarray*}
\mathbb{E} \left[ \left\|g^k\right\|^2\;|\;{\bf J}^k, x^k \right] &\leq & 
 2\mathbb{E} \left[  \left\| ({\bf G}(x^k) - {\bf G}(x^*)){\theta}_{S_k}{\bf \Pi}_{{\bf I}_{S_k}}\lambda \right\|^2\;|\;{\bf J}^k, x^k  \right] \\
 && \qquad + 2 \mathbb{E} \left[ \left\| ({\bf J}^k - {\bf G}(x^*)){\theta}_{S_k}{\bf \Pi}_{{\bf I}_{S_k}}\lambda - {\bf J}^k \lambda \right\|^2 \;|\; {\bf J}^k, x^k  \right].
\end{eqnarray*}

\noindent For $\mathbb{E} \left[  \left\| ({\bf G}(x^k) - {\bf G}(x^*)){\theta}_{S_k}{\bf \Pi}_{{\bf I}_{S_k}}\lambda \right\|^2\;|\;{\bf J}^k, x^k  \right]$, by applying Lemma \ref{lm:ineq} with $z = |C|$, we get 
\begin{eqnarray*}
\mathbb{E} \left[  \left\| ({\bf G}(x^k) - {\bf G}(x^*)){\theta}_{S_k}{\bf \Pi}_{{\bf I}_{S_k}}\lambda \right\|^2\;|\;{\bf J}^k, x^k  \right] &=& \sum_C p_C \left\| ({\bf G}(x^k) - {\bf G}(x^*)){\theta}_{C}{\bf \Pi}_{{\bf I}_{C}}\lambda \right\|^2 \\
&\leq& \sum_C p_C |C| \sum_{i\in C} \left\|{\theta}_C^i \lambda_i (\nabla f_i(x^k) - \nabla f_i(x^*))\right\|^2 \\
&=& \sum_{i=1}^n \lambda_i^2\sum_{C: i\in C} p_C |C| ({\theta}_C^i)^2 \left\|\nabla f_i(x^k) - \nabla f_i(x^*)\right\|^2 \\
&=& \sum_{i=1}^n \beta_i \lambda_i^2 \left\|\nabla f_i(x^k) - \nabla f_i(x^*) \right\|^2.
\end{eqnarray*}

\noindent For $\mathbb{E} \left[ \left\| ({\bf J}^k - {\bf G}(x^*)){\theta}_{S_k}{\bf \Pi}_{{\bf I}_{S_k}}\lambda - {\bf J}^k \lambda \right\|^2\;|\;{\bf J}^k, x^k  \right]$, since 
$$
\mathbb{E} \left[ ({\bf J}^k - {\bf G}(x^*)){\theta}_{S_k}{\bf \Pi}_{{\bf I}_{S_k}}\lambda\;|\;{\bf J}^k, x^k  \right] = {\bf J}^k \lambda,
$$
and $\mathbb{E}[\|X - \mathbb{E}[X]\|^2] \leq \mathbb{E}[\|X\|^2]$, we have that 
\begin{eqnarray*}
\mathbb{E} \left[ \left\| ({\bf J}^k - {\bf G}(x^*)){\theta}_{S_k}{\bf \Pi}_{{\bf I}_{S_k}}\lambda - {\bf J}^k \lambda \right\|^2\;|\;{\bf J}^k, x^k  \right] &\leq& \mathbb{E} \left[ \left\| ({\bf J}^k - {\bf G}(x^*)){\theta}_{S_k}{\bf \Pi}_{{\bf I}_{S_k}}\lambda  \right\|^2\;|\;{\bf J}^k, x^k  \right] \\
&=& \sum_{C} p_C \left\| {\bf J}^k - {\bf G}(x^*)){\theta}_{C}{\bf \Pi}_{{\bf I}_{C}}\lambda \right\|^2 \\
&\leq& \sum_{C} p_C |C| \sum_{i\in C} \left\|{\theta}_C^i\lambda_i ({\bf J}^k_{:i} - \nabla f_i(x^*))\right\|^2 \\
&=& \mathbb{E}\left[ |S|\sum_{i\in S} \left\|{\theta}_S^i\lambda_i ({\bf J}^{k}_{:i} - \nabla f_i(x^*))\right\|^2\;|\;{\bf J}^k, x^k\right],
\end{eqnarray*}
where the second inequality comes from Lemma \ref{lm:ineq} with $z=|C|$.

Finally, we arrive at the result 
$$
\mathbb{E}\left[ \|g^k\|^2\;|\;{\bf J}^k, x^k \right] 
\leq  2\mathbb{E}\left[ |S|\sum_{i\in S} \left\|{\theta}_S^i\lambda_i ({\bf J}^{k}_{:i} - \nabla f_i(x^*))\right\|^2\;|\;{\bf J}^k, x^k \right] 
+ 2\sum_{i=1}^n \beta_i\lambda_i^2 \left\|\nabla f_i(x^k) - \nabla f_i(x^*) \right\|^2.
$$
\end{proof}

\begin{lemma}\label{lm:nablafx}
If $f$ is $\mu$-strongly convex, and $f_i$ is convex and $L_i$-smooth, then 
\[
\langle \nabla f(x) - \nabla f(y), x-y \rangle  \geq \frac{\mu}{2}\| x-y\|^2 + \sum_{i=1}^n \frac{\lambda_i}{2L_i} \left\|\nabla f_i(x) - \nabla f_i(y)\right\|^2,
\]
for all $x$, $y \in \mathbb{R}^d$. 
\end{lemma}

\begin{proof}
Since $f$ is $\mu$-strongly convex, we have 
\begin{equation}\label{fmustrongly}
\langle \nabla f(x) - \nabla f(y), x -y \rangle \geq \mu \|x-y\|^2.
\end{equation}

Since $f_i$ is $L_i$-smooth, we have 
$$
\langle \nabla f_i(x) - \nabla f_i(y), x-y \rangle \geq \frac{1}{L_i} \| \nabla f_i(x) - \nabla f_i(y) \|^2,
$$
which indicates 
\begin{equation}\label{Lsmooth}
\langle \nabla f(x) - \nabla f(y), x-y \rangle \geq \sum_{i=1}^n \frac{\lambda_i}{L_i} \| \nabla f_i(x) - \nabla f_i(y) \|^2.
\end{equation}

\noindent Combining (\ref{fmustrongly}) and (\ref{Lsmooth}), we can get 
$$
\langle \nabla f(x) - \nabla f(y), x-y \rangle  \geq \frac{\mu}{2}\| x-y\|^2 + \sum_{i=1}^n \frac{\lambda_i}{2L_i} \|\nabla f_i(x) - \nabla f_i(y)\|^2.
$$
\end{proof}

\subsection{Proof of Theorem~\ref{Th:convp}}

Having established the above lemmas, we are ready to proceed to the proof of pour main theorem covering the smooth case (Theorem~\ref{Th:convp}).

Let $\mathbb{E}_k[\cdot]$ denote the expectation conditional on ${\bf J}^k$ and $x^k$. First, from Assumption \ref{ass:thetaS}, it is evident that 
\begin{equation}\label{expgk}
\mathbb{E}_k[g^k] = {\bf J}^k \lambda + \nabla f(x^k) - {\bf J}^k \lambda = \nabla f(x^k). 
\end{equation}
Then we can obtain  
\begin{eqnarray*}
\mathbb{E}_k\left[ \|x^{k+1} - x^*\|^2 \right] &=& \mathbb{E}_k\left[ \|x^k - x^* - \alpha g^k\|^2 \right] \\
&\overset{(\ref{expgk})}{=}& \| x^k -x^*\|^2 - 2\alpha \langle \nabla f(x^k), x^k-x^* \rangle + \alpha^2 \mathbb{E}_k[\| g^k \|^2] \\
&\overset{\rm Lemma~\ref{lm:nablafx}}{\leq}&  (1-\mu \alpha)\|x^k - x^*\|^2 - 2\alpha \sum_{i=1}^n \frac{\lambda_i}{2L_i} \|\nabla f_i(x^k) - \nabla f_i(x^*) \|^2 + \alpha^2 \mathbb{E}_k[\|g^k\|^2] \\
&\overset{\rm Lemma~\ref{lm:gk2}}{\leq}& (1-\mu \alpha)\|x^k-x^*\|^2 + 2\alpha \sum_{i=1}^n \left[ (\alpha \beta_i \lambda_i^2 - \frac{\lambda_i}{2L_i})\|\nabla f_i(x^k) - \nabla f_i(x^*)\|^2 \right] \\
&& + 2\alpha^2 \mathbb{E}_k\left[ |S|\sum_{i\in S} \left\|{\theta}_S^i\lambda_i ({\bf J}^{k}_{:i} - \nabla f_i(x^*))\right\|^2 \right] 
\end{eqnarray*}
Taking expectation again and applying the tower property, we obtain  
\begin{eqnarray*}
\mathbb{E}\left[ \|x^{k+1} - x^*\|^2 \right] &\leq& (1-\mu \alpha) \mathbb{E}\left[ \|x^k-x^*\|^2 \right] + 2\alpha \mathbb{E} \left[ \sum_{i=1}^n \left( (\alpha \beta_i \lambda_i^2 - \frac{\lambda_i}{2L_i})\left\|\nabla f_i(x^k) - \nabla f_i(x^*)\right\|^2 \right) \right]\\
&& + 2\alpha^2 \mathbb{E}\left[ |S|\sum_{i\in S} \left\|{\theta}_S^i\lambda_i ({\bf J}^{k}_{:i} - \nabla f_i(x^*))\right\|^2 \right]. 
\end{eqnarray*}
Therefore, for the stochastic Lyapunov function $\Psi^{k+1}_S$, we have 
\begin{eqnarray*}
\mathbb{E}[\Psi^{k+1}_S] &\leq& (1-\mu \alpha) \mathbb{E}\left[ \|x^k-x^*\|^2 \right] + 2\alpha \mathbb{E} \left[ \sum_{i=1}^n \left( \left(\alpha \beta_i \lambda_i^2 - \frac{\lambda_i}{2L_i}\right)\left\|\nabla f_i(x^k) - \nabla f_i(x^*)\right\|^2 \right) \right]\\
&& + 2\alpha^2 \mathbb{E}\left[ |S|\sum_{i\in S} \left\|{\theta}_S^i\lambda_i ({\bf J}^{k}_{:i} - \nabla f_i(x^*))\right\|^2 \right] + 2\alpha \mathbb{E}\left[ |S|\sum_{i\in S}\sigma_i \left\|{\theta}_S^i\lambda_i ({\bf J}^{k+1}_{:i} - \nabla f_i(x^*))\right\|^2 \right] \\
&\overset{\rm Lemma \ref{lm:Jk+1}}{\leq}& \mathbb{E}\left[ (1-\mu \alpha)\|x^k-x^*\|^2 \right] + 2\alpha \mathbb{E} \left[ \sum_{i=1}^n \left( (\alpha \beta_i \lambda_i^2 - \frac{\lambda_i}{2L_i} + \sigma_i p_i\beta_i\lambda_i^2)\left\|\nabla f_i(x^k) - \nabla f_i(x^*)\right\|^2 \right) \right]\\
&& + 2\alpha \mathbb{E}\left[ |S|\sum_{i\in S} [(1-p_i)\sigma_i + \alpha]\left\|{\theta}_S^i\lambda_i ({\bf J}^{k}_{:i} - \nabla f_i(x^*))\right\|^2 \right] 
\end{eqnarray*}

In order to guarantee that $\mathbb{E}{\Psi^{k+1}_S} \leq (1-\mu \alpha)\mathbb{E}{\Psi^k_S}$, $\alpha$ should be chosen such that 
$$
\alpha \beta_i \lambda_i^2 - \frac{\lambda_i}{2L_i} + \sigma_i p_i\beta_i\lambda_i^2 \leq 0, \quad \Rightarrow \quad \alpha \leq \frac{1}{2L_i\beta_i\lambda_i} - \sigma_ip_i, 
$$
and 
$$
(1-p_i)\sigma_i + \alpha \leq (1- \mu \alpha)\sigma_i, \quad \Rightarrow \quad \alpha \leq \frac{\sigma_ip_i}{\mu\sigma_i + 1},
$$
for all $1\leq i\leq n$. Since $\sigma_i = \frac{1}{4L_i\beta_ip_i\lambda_i}$, if $\alpha$ satisfies
$$
\alpha \leq \min_{1\leq i\leq n} \frac{p_i}{\mu + 4L_i\beta_i\lambda_ip_i},
$$
then we have the recursion $\mathbb{E}{\Psi^{k+1}_S} \leq (1-\mu \alpha)\mathbb{E}{\Psi^k_S}$.

\subsection{Proof of Theorem \ref{Th:convps}}

In problem (\ref{problemn}), 
$$
f(x) = \sum_{C\in {\cal G}}\frac{|C|}{n} f_C(x) = \sum_{i=1}^m \frac{|C_i|}{n} f_{C_i}(x).
$$
Hence by choosing ${\tilde f}_i = f_{C_i}$ and ${\tilde \lambda}_i = \frac{|C_i|}{n}$ for $1\leq i\leq m$, problem (\ref{problemn}) has the same form as problem (\ref{primal}), and the Lipschitz smothness constant ${\tilde L}_i$ of ${\tilde f}_i$ is $L_{C_i}$.  

\vskip 2mm

From the partition sampling $S$, we can construct a sampling $\tilde S$ from $S$ as follows: ${\tilde S}(S) = \{ i \}$ if $S = C_i$. It is obvious that  $\mathbb{P}[{\tilde S} =\{i \}] = p_{C_i}$ for all $1\leq i\leq m$. For sampling $\tilde S$, we choose ${\theta}_{\tilde S}$ be such that ${\theta}_{\{i \}} = \frac{1}{p_{C_i}} {\bf I}$. For the sampling $\tilde S$ and ${\theta}_{\tilde S}$, the corresponding ${\tilde \beta}_i = \frac{1}{p_{C_i}}$ from (\ref{betai}). 

\vskip 2mm

Let ${\tilde S}_k(S_k) = \{ i \}$ if $S_k = C_i$ for $k\geq 0$. Then ${\tilde S}_k \sim {\tilde S}$ is equivalent to $S_k \sim S$.  Let $\{ {\tilde x}^k, {\tilde {\bf J}}^k \}$ be produced by Algorithm \ref{minibatchSAGA} with ${\tilde \lambda}_i$, ${\tilde f}_i$, $\tilde S$, and ${\theta}_{\tilde S}$. Then it is easy to see that 
\begin{equation}
\left\{
\begin{array}{l}
{\tilde x}^k = x^k,\\
{\tilde {\bf J}}^k_{:i} = \frac{1}{|C_i|}{\bf J}^k{\Pi}_{{\bf I}_{C_i}}e, \ \forall 1\leq i\leq m.
\end{array}\right.
\end{equation}

Therefore, by applying Theorem~\ref{Th:convp} to $\{{\tilde x}^k,  {\tilde {\bf J}}^k\}$, we can get the results.

\newpage
\section{Nonsmooth Case: Proofs of Theorems~\ref{th:phi} and~\ref{th:phistrongly}}

\subsection{Lemmas}

\begin{lemma}\label{l:gkexp}
	Let $\mathbb{E}_k[\cdot]$ denote the expectation conditional on ${\bf J}^k$ and $x^k$. First, from Assumption \ref{ass:thetaS}, it is evident that 
	\begin{equation}\label{expgk2}
	\mathbb{E}_k[g^k] = {\bf J}^k \lambda + \nabla f(x^k) - {\bf J}^k \lambda = \nabla f(x^k). 
	\end{equation}
\end{lemma}

\begin{lemma}\label{l:nadph} Under Assumption~\ref{ass:phi} and Assumption~\ref{ass:uniquegradient}, for any $x^*, y^*\in \cX^*$ and $x\in \R^d$, we have,
	$$
	\<\nabla f(x)-\nabla f(y^*), x-y^*> \geq \gamma \sum_{i=1}^{n} \lambda_i\left\| \nabla\phi_i({\bf A}_i^\top x)-\nabla \phi_i({\bf A}_i^\top x^*)\right\|^2.
	$$
\end{lemma}
\begin{proof}
	Since $\phi_i$ is $1/\gamma$-smooth, we have 
	$$
	\langle \nabla \phi_i(\tilde x) - \nabla \phi_i(\tilde y), {\tilde x}-{\tilde y}\rangle \geq \gamma \left\| \nabla \phi_i(\tilde x) - \nabla \phi_i(\tilde y)\right\|^2, \enspace \forall i=1,\cdots,n.
	$$
	Let $\tilde x = {\bf A}_i^\top x$, and $\tilde y = {\bf A}_i^\top y^*$ in the above inequlity. Then we get 
	$$
	\langle \nabla \phi_i({\bf A}_i^\top x) - \nabla \phi_i({\bf A}_i^\top y^*), {\bf A}_i(x-y^*)\rangle \geq \gamma \left\| \nabla \phi_i({\bf A}_i^\top x) - \nabla \phi_i({\bf A}_i^\top y^*)\right\|^2, \enspace \forall i=1,\cdots,n,
	$$
	which is actually 
	$$
	\langle \nabla f_i(x)-\nabla f_i(y^*), x-y^*\rangle \geq \gamma \left\| \nabla\phi_i({\bf A}_i^\top x)-\nabla \phi_i({\bf A}_i^\top y^*)\right\|^2, \enspace \forall i=1,\cdots,n,
	$$
	
	Summing over $i$ we get
	$$
	\langle \nabla f(x)-\nabla f(y^*), x-y^* \rangle \geq \gamma \sum_{i=1}^{n} \lambda_i\left\| \nabla\phi_i({\bf A}_i^\top x)-\nabla \phi_i({\bf A}_i^\top y^*)\right\|^2
	$$
	
	The statement then follows directly from Assumption~\ref{ass:uniquegradient}.
\end{proof}

\iffalse
\begin{lemma}\label{l:nadph2}
	There is $\mu>0$ such that for any $x^*\in \cX^*$,
	$$
	\<\nabla f(x), x-x^*>\geq {\mu}\|x- [x]^*\|^2.
	$$
\end{lemma}
\begin{proof}
	By Theorem 9 in~\cite{NecoaraNesGli}, we know that there is some $\mu>0$ such that,
	$$
	f(x)-f^*\geq  \mu \|x- [x]^*\|^2,\enspace \forall x\in\R^d.
	$$
	Then the statement holds as $f(x^*)\geq f(x)+\<\nabla f(x),x^*-x>$.
\end{proof}

\fi
We now quote a result from~\cite{pmlr-v80-poon18a}.
\begin{theorem}\label{th:poon18}[\cite{pmlr-v80-poon18a}]
If  if $f$ is $L$-smooth and $\alpha\leq \frac{1}{3L}$, then almost surely the sequence $P(x^k)$ is bounded.
\end{theorem}

%\peter{Is this result from \cite{pmlr-v80-poon18a} or \cite{prox-SVRG}? We mention both, which is confusing.}
\begin{lemma}\label{l:prox-svrg}
	Under Assumption~\ref{ass:rsc}, if $f$ is $L$-smooth and $\alpha\leq \frac{1}{3L}$, then there is $\mu>0$ such that for all $k\geq 0$,
	$$\Exp_k\left[ \|x^{k+1}-[x^{k+1}]^*\|^2 \right]
	\leq \frac{1}{1+\mu \alpha} \Exp_k\left[\|x^k-[x^k]^*\|^2\right]+\frac{2\alpha^2}{1+\mu\alpha} \Exp_k\left[\|g^k-\nabla f(x^k)\|^2\right]$$
	\footnote{This result was mainly proved in~\cite{prox-SVRG}. For completeness, we include a proof. } 
\end{lemma}

\begin{proof}
	Since $$
	x^{k+1}=\arg\min\left\{\frac{1}{2\alpha}\|y-x^k+\alpha g^k\|^2+\psi(y)\right\},
	$$
	we know $\alpha^{-1}(x^k-x^{k+1})-g^k\in \partial \psi(x^{k+1})$.
	Using the convexity of $f$ and $\psi$, we obtain
	\begin{align}\label{alp:2}
	P^*\geq f(x^k)+\<\nabla f(x^k),[x^k]^*-x^k>+\psi(x^{k+1})+\< \alpha^{-1}(x^k-x^{k+1})-g^k,[x^k]^*-x^{k+1} >.
	\end{align}
	Next we bound $f(x^{k+1})$ by $f(x^k)$. Since $f$ is $L$-smooth, we have 
	\begin{align}\label{alp:1}
	f(x^{k+1})\leq f(x^k)+\<\nabla f(x^k), x^{k+1}-x^k>+ \frac{ L}{2}\|x^{k+1}-x^k\|^2.
	\end{align}
	Combining~\eqref{alp:2} and~\eqref{alp:1} we get
	\begin{eqnarray*}
		f(x^{k+1})+\psi(x^{k+1})-P^*&\leq& \<\nabla f(x^k), x^{k+1}-[x^k]^*>+\frac{ L}{2}\|x^{k+1}-x^k\|^2\\
		&&\qquad\quad -\< \alpha^{-1}(x^k-x^{k+1})-g^k,[x^k]^*-x^{k+1} >\\
		&=& \frac{ L}{2}\|x^{k+1}-x^k\|^2 -\frac{1}{\alpha}\<x^k-x^{k+1},[x^k]^*-x^{k+1} >+\<\nabla f(x^k)-g^k,[x^k]^*-x^{k+1} >\\
		&=& \left(\frac{ L}{2}-\frac{1}{2\alpha}\right)\|x^{k+1}-x^k\|^2 -\frac{1}{2\alpha}\|x^{k+1}-x^k\|^2\\&&\qquad-\frac{1}{\alpha}\<x^k-x^{k+1},[x^k]^*-x^k> +\<\nabla f(x^k)-g^k,[x^k]^*-x^{k+1} >
		\\&\leq & -\frac{1}{2\alpha}\|x^{k+1}-x^k\|^2-\frac{1}{\alpha}\<x^k-x^{k+1},[x^k]^*-x^k> +\<\nabla f(x^k)-g^k,[x^k]^*-x^{k+1} >
		\\&=& -\frac{1}{2\alpha} \|x^{k+1}-[x^k]^*\|^2+\frac{1}{2\alpha} \|x^{k}-[x^k]^*\|^2+\<\nabla f(x^k)-g^k,[x^k]^*-x^{k+1} >.
	\end{eqnarray*}
	Now we use Assumption~\ref{ass:rsc} and Theorem~\ref{th:poon18} to obtain the existence of $\mu>0$ such that
	$$
	{(\mu\alpha+1)}\|x^{k+1}-[x^{k+1}]^*\|^2\leq  \|x^{k}-[x^k]^*\|^2+2\alpha \<\nabla f(x^k)-g^k,[x^k]^*-x^{k+1} >.
	$$
	Taking conditional expectation on both side, we get
	\begin{eqnarray*}
		&&\Exp_k\left[\|x^{k+1}-[x^{k+1}]^*\|^2 \right]\leq  \frac{1}{\mu\alpha+1}\Exp_k\left[ \|x^{k}-[x^{k}]^*\|^2\right]+\frac{2\alpha}{\mu\alpha+1}\Exp_k\left[\<\nabla f(x^k)-g^k,[x^k]^*-x^{k+1} >\right]\\
		&&\overset{\eqref{expgk2}}{=}\frac{1}{\mu\alpha+1}\Exp_k\left[ \|x^{k}-[x^{k}]^*\|^2 \right]+\frac{2\alpha}{\mu\alpha+1}\Exp_k\left[\<\nabla f(x^k)-g^k,\bar{x}^{k+1}-x^{k+1} >\right],
	\end{eqnarray*}
	where $\bar{x}^{k+1}=\prox^\psi_{\alpha}(x^k-\alpha \nabla f(x^k))$ is determined by $x^k$. Recall that  ${x}^{k+1}=\prox^\psi_{\alpha}(x^k-\alpha g^k)$.
	We use the non-expansiveness of the proximal mapping:
	$$
	\| \bar{x}^{k+1}-x^{k+1} \|\leq \alpha \|\nabla f(x^k)-g^k\|.
	$$
	Then the statement follows by the  Cauchy-Schwarz inequality.
\end{proof}

\begin{lemma}\label{lm:ak+1}
	\begin{eqnarray*}
		\mathbb{E} \left[  \sum_{i=1}^n\sigma_i  p_i^{-1}v_i\lambda_i^2 \left\|\alpha_i^{k+1}-\nabla \phi_i({\bf A}_i^\top x^*)\right\|^2 \right]&\leq&\mathbb{E} \left[  \sum_{i=1}^n \sigma_i  v_i\lambda_i^2 \left\|\nabla \phi({\bf A}_i^\top x^k)-\nabla \phi_i({\bf A}_i^\top x^*)\right\|^2  \right] \\&&\qquad+ \mathbb{E} \left[  \sum_{i=1}^n \sigma_i  p_i^{-1} v_i\lambda_i^2(1-p_i)\left\|\alpha_i^{k}-\nabla \phi_i({\bf A}_i^\top x^*)\right\|^2 \right]
	\end{eqnarray*}
\end{lemma}

\begin{proof}
	First notice that 
	$$
	\alpha^{k+1}_{i} = \left\{ \begin{array}{ll}
	\nabla \phi({\bf A}_i^\top x^k),& \mbox{ if $i \in { S^k}$} \\
	\alpha_i^k ,&\mbox{ if $i \notin S^k$ }
	\end{array} \right.
	$$
	Then by taking conditional expectation on $\alpha^k$, we have 
	\begin{eqnarray*}
		&&\mathbb{E} \left[  \sum_{i=1}^n\sigma_i  p_i^{-1}v_i\lambda_i^2 \left\|\alpha_i^{k+1}-\nabla \phi_i({\bf A}_i^\top x^*)\right\|^2 \;|\;\alpha^k \right] \\
		&=&  \sum_{i=1}^n \sigma_i  p_i^{-1}v_i\lambda_i^2 \Exp\left[\left\|\alpha_i^{k+1}-\nabla \phi_i({\bf A}_i^\top x^*)\right\|^2  |\alpha^k \right]\\
		&=&  \sum_{i=1}^n \sigma_i  p_i^{-1}v_i\lambda_i^2 \left(p_i\|\nabla \phi_i({\bf A}_i^\top x^k)-\nabla \phi_i({\bf A}_i^\top x^*)\|^2  +(1-p_i)\left\|\alpha_i^{k}-\nabla \phi_i({\bf A}_i^\top x^*)\right\|^2 \right).
		\end {eqnarray*}
		
		Taking expectations again and applying the tower property, we obtain the result. 
		
	\end{proof}
	
	\begin{lemma}\label{lm:gk23}
		For any $x^*,y^*\in \cX^*$,
		$$\mathbb{E}\left[ \|g^k-\nabla f(y^*)\|^2\;|\; {\bf J}^k, x^k\right] \leq
		2\sum_{i=1}^n p_i^{-1}v_i \lambda_i^2\left\|\nabla \phi_i({\bf A}_i^\top x^k)-\nabla \phi_i({\bf A}_i^\top x^*)\right\|^2 +2\sum_{i=1}^n p_i^{-1}v_i \lambda_i^2\left\|\alpha_i^k-\nabla \phi_i({\bf A}_i^\top x^*)\right\|^2 .
		$$
	\end{lemma}
	
	\begin{proof}
		Recall that 
		\begin{eqnarray*}
			g^k &=& {\bf J}^k \lambda + \left({\bf G}(x^k) - {\bf J}^k\right){\theta}_{S_k}{\bf \Pi}_{{\bf I}_{S_k}} \lambda \\
			&=& {\bf J}^k \lambda - \left({\bf J}^k - {\bf G}(x^*)\right){\theta}_{S_k}{\bf \Pi}_{{\bf I}_{S_k}}\lambda + ({\bf G}(x^k) - {\bf G}(x^*)){\theta}_{S_k}{\bf \Pi}_{{\bf I}_{S_k}}\lambda.
		\end{eqnarray*}
		By applying Lemma \ref{lm:ineq} with $z=2$, we get 
		$$
		\|g^k-\nabla f(y^*)\|^2 \leq 2 \left\| ({\bf G}(x^k) - {\bf G}(x^*)){\theta}_{S_k}{\bf \Pi}_{{\bf I}_{S_k}}\lambda \right\|^2 + 2 \left\| ({\bf J}^k - {\bf G}(x^*)){\theta}_{S_k}{\bf \Pi}_{{\bf I}_{S_k}}\lambda - {\bf J}^k \lambda +\nabla f(y^*) \right\|^2, 
		$$
		which implies that 
		\begin{eqnarray*}
			\mathbb{E}\left[ \|g^k-\nabla f(y^*)\|^2\;|\; {\bf J}^k, x^k\right] &\leq & 
			2\mathbb{E} \left[  \left\| ({\bf G}(x^k) - {\bf G}(x^*)){\theta}_{S_k}{\bf \Pi}_{{\bf I}_{S_k}}\lambda \right\|^2\;|\;{\bf J}^k, x^k  \right] \\
			&& \qquad + 2 \mathbb{E} \left[ \left\| ({\bf J}^k - {\bf G}(x^*)){\theta}_{S_k}{\bf \Pi}_{{\bf I}_{S_k}}\lambda - {\bf J}^k \lambda +\nabla f(y^*)\right\|^2 \;|\; {\bf J}^k, x^k  \right]\\
			&\leq & 2\mathbb{E} \left[ \left \| ({\bf G}(x^k) - {\bf G}(x^*)){\theta}_{S_k}{\bf \Pi}_{{\bf I}_{S_k}}\lambda  \right\|^2\;|\;{\bf J}^k, x^k  \right] \\
			&& \qquad + 2 \mathbb{E} \left[ \left\| ({\bf J}^k - {\bf G}(x^*)){\theta}_{S_k}{\bf \Pi}_{{\bf I}_{S_k}}\lambda  \right\|^2\;|\;{\bf J}^k, x^k  \right] ,
		\end{eqnarray*}
		where the last inequality follows from $$
		\mathbb{E} \left[ ({\bf J}^k - {\bf G}(x^*)){\theta}_{S_k}{\bf \Pi}_{{\bf I}_{S_k}}\lambda\;|\;{\bf J}^k, x^k  \right] = {\bf J}^k \lambda-\nabla f(x^*)\overset{Assumption~\ref{ass:uniquegradient}}{=} {\bf J}^k \lambda-\nabla f(y^*)
		$$
		and $\mathbb{E}[\|X - \mathbb{E}[X]\|^2] \leq \mathbb{E}[\|X\|^2]$.
		Hence,
		\begin{eqnarray*}
			\mathbb{E}\left[ \|g^k-\nabla f(x^*)\|^2\;|\; {\bf J}^k, x^k\right]  &\leq &2
			\Exp \left[\left\|\sum_{i\in S_k} p_i^{-1}\lambda_i{\bf A}_i\left(\phi_i'({\bf A}_i^\top x^k)-\phi'_i({\bf A}_i^\top x^*)\right)\right\|^2 \;|\; {\bf J}^k, x^k\right]\\&&\qquad+
			2\Exp\left[\left\|\sum_{i\in S_k} p_i^{-1}\lambda_i{\bf A}_i\left(\alpha^k-\phi'_i({\bf A}_i^\top x^*)\right)\right\|^2\;|\; {\bf J}^k, x^k \right]\\
			\\ &\overset{\eqref{eq:ESOfirst}}{\leq} & 2\sum_{i=1}^n p_i^{-1}v_i \lambda_i^2\left\|\phi_i'({\bf A}_i^\top x^k)-\phi'_i({\bf A}_i^\top x^*)\right\|^2 +2\sum_{i=1}^n p_i^{-1}v_i \lambda_i^2\left\|\alpha_i^k-\phi'_i({\bf A}_i^\top x^*)\right\|^2 .
		\end{eqnarray*}

	\end{proof}
	
	\begin{lemma}\label{lm:gk233}
		For any $x^*\in \cX^*$,
		$$\mathbb{E}\left[ \|g^k-\nabla f(x^k)\|^2\;|\; {\bf J}^k, x^k\right] \leq
		2\sum_{i=1}^n p_i^{-1}v_i \lambda_i^2\left\|\nabla \phi_i({\bf A}_i^\top x^k)-\nabla \phi_i({\bf A}_i^\top x^*)\right\|^2 +2\sum_{i=1}^n p_i^{-1}v_i \lambda_i^2\left\|\alpha_i^k-\nabla \phi_i({\bf A}_i^\top x^*)\right\|^2 .
		$$
	\end{lemma}
	\begin{proof}
		\begin{eqnarray*}
			g^k -\nabla f(x^k)
			&=& {\bf J}^k \lambda - ({\bf J}^k - {\bf G}(x^*)){\theta}_{S_k}{\bf \Pi}_{{\bf I}_{S_k}}\lambda + ({\bf G}(x^k) - {\bf G}(x^*)){\theta}_{S_k}{\bf \Pi}_{{\bf I}_{S_k}}\lambda-\nabla f(x^k)\\
			&=& {\bf J}^k \lambda - ({\bf J}^k - {\bf G}(x^*)){\theta}_{S_k}{\bf \Pi}_{{\bf I}_{S_k}}\lambda -\nabla f(x^*)
			\\&&\qquad\quad+ ({\bf G}(x^k) - {\bf G}(x^*)){\theta}_{S_k}{\bf \Pi}_{{\bf I}_{S_k}}\lambda-\nabla f(x^k)+\nabla f(x^*).
		\end{eqnarray*}
		The rest of the proof is the same as in Lemma~\ref{lm:gk23}.
	\end{proof}
	
	\subsection{Proof of Theorem \ref{th:phi}}

		For any $x^*\in \cX^*$, we have
		$$
		x^*=\prox^\psi_{\alpha}( x^*-\alpha \nabla f(x^*)).
		$$
		Therefore,
		\begin{eqnarray*}
			&&\mathbb{E}_k\left[ \left\|x^{k+1} - [x^{k+1}]^*\right\|^2 \right] \\&\leq&
			\mathbb{E}_k\left[ \left\|x^{k+1} - [x^{k}]^*\right\|^2 \right]\\ &=&
			\mathbb{E}_k\left[ \left\| \prox^\psi_{\alpha}(x^k  - \alpha g^k)-[x^k]^*\right\|^2 \right] \\
			&=&\mathbb{E}_k\left[ \left\| \prox^\psi_{\alpha}(x^k  - \alpha g^k)-\prox^\psi_{\alpha}([x^k]^*-\alpha \nabla f([x^k]^*)) \right\|^2\right]\\
			&\leq & \mathbb{E}_k\left[ \left\| x^k  - \alpha g^k-([x^k]^*-\alpha \nabla f([x^k]^*)) \right\|^2\right]\\
			\qquad &\overset{(\ref{expgk2})}{=}& \left\| x^k -[x^k]^*\right\|^2 - 2\alpha \langle \nabla f(x^k)-\nabla f([x^k]^*), x^k-[x^k]^* \rangle + \alpha^2 \mathbb{E}_k\left[\| g^k-\nabla f([x^k]^*) \|^2\right] \\
		\end{eqnarray*}
		Now we apply Lemma~\ref{l:prox-svrg}.  For any $\beta\in [0,1]$, we have
		\begin{eqnarray*}
			&&\mathbb{E}_k\left[ \left\|x^{k+1} - [x^{k+1}]^*\right\|^2 \right] \\&\leq&
			\left( \frac{\beta}{1+\alpha\mu}+1-\beta\right)\left\| x^k -[x^k]^*\right\|^2 -2\alpha(1-\beta)\langle \nabla f(x^k)-\nabla f([x^k]^*), x^k-[x^k]^* \rangle\\
			&&\qquad+ \alpha^2(1-\beta) \mathbb{E}_k\left[\| g^k-\nabla f([x^k]^*) \|^2\right] +\frac{2\alpha^2\beta}{1+\mu\alpha} \Exp_k\left[\|g^k-\nabla f(x^k)\|^2\right]
		\end{eqnarray*}
		Plugging in Lemma~\ref{l:nadph}, Lemma~\ref{lm:gk23} and Lemma~\ref{lm:gk233} we obtain:
		\begin{eqnarray*}
			&&\mathbb{E}_k\left[ \left\|x^{k+1} - [x^{k+1}]^*\right\|^2 \right] \\&\leq&
			\left( \frac{\beta}{1+\alpha\mu}+1-\beta\right)\left\| x^k -[x^k]^*\right\|^2 -2\gamma\alpha(1-\beta)\sum_{i=1}^{n} \lambda_i\left\| \nabla \phi_i({\bf A}_i^\top x^k)-\nabla \phi_i({\bf A}_i^\top x^*)\right\|^2\\
			&&\quad+  \left(\alpha^2(1-\beta)+\frac{2\alpha^2\beta}{1+\mu\alpha} \right)\left(2\sum_{i=1}^n p_i^{-1}v_i \lambda_i^2\left\|\nabla \phi_i({\bf A}_i^\top x^k)-\nabla \phi_i({\bf A}_i^\top x^*)\right\|^2 +2\sum_{i=1}^n p_i^{-1}v_i \lambda_i^2\left\|\alpha_i^k-\nabla \phi_i({\bf A}_i^\top x^*)\right\|^2 \right)\\
			&=&\left( \frac{\beta}{1+\alpha\mu}+1-\beta\right)\left\| x^k  -[x^k]^*\right\|^2+\sum_{i=1}^n 2p_i^{-1}v_i \lambda_i^2\left(\alpha^2(1-\beta)+\frac{2\alpha^2\beta}{1+\mu\alpha} \right)\left\|\alpha_i^k-\nabla \phi_i({\bf A}_i^\top x^*)\right\|^2  \\&&\qquad -\sum_{i=1}^{n}\left( 2\gamma\lambda_i \alpha(1-\beta)- 2p_i^{-1}v_i \lambda_i^2\left(\alpha^2(1-\beta)+\frac{2\alpha^2\beta}{1+\mu\alpha} \right)\right) \left\|\nabla \phi_i({\bf A}_i^\top x^k)-\nabla \phi_i({\bf A}_i^\top x^*)\right\|^2\\
		\end{eqnarray*}
		Taking expectation again and applying the tower property, we obtain  
		\begin{eqnarray*}
			\mathbb{E}\left[ \|x^{k+1} - [x^{k+1}]^*\|^2 \right] &\leq&  \left( \frac{\beta}{1+\alpha\mu}+1-\beta\right)\mathbb{E}\left[ \left\| x^k  -[x^k]^*\right\|^2\right]\\&&\qquad+\sum_{i=1}^n 2p_i^{-1}v_i \lambda_i^2\left(\alpha^2(1-\beta)+\frac{2\alpha^2\beta}{1+\mu\alpha} \right)\Exp\left[\left\|\alpha_i^k-\nabla \phi_i({\bf A}_i^\top x^*)\right\|^2 \right] \\&&\qquad -\sum_{i=1}^{n}\left( 2\gamma \lambda_i \alpha(1-\beta)- 2p_i^{-1}v_i \lambda_i^2\left(\alpha^2(1-\beta)+\frac{2\alpha^2\beta}{1+\mu\alpha} \right)\right) \Exp\left[\left\|\nabla \phi_i({\bf A}_i^\top x^k)-\nabla \phi_i({\bf A}_i^\top x^*)\right\|^2\right]\end{eqnarray*}
		Therefore, for the stochastic Lyapunov function $\Psi^{k+1}$, we have in view of Lemma~\ref{lm:ak+1},
		\begin{eqnarray*}
			\mathbb{E}\left[ \Psi^{k+1}\right] &\leq&  \left( \frac{\beta}{1+\alpha\mu}+1-\beta\right)\mathbb{E}\left[ \left\| x^k  -[x^k]^*\right\|^2\right]\\&&\qquad+\sum_{i=1}^n 2p_i^{-1}v_i \lambda_i^2\left(\alpha^2(1-\beta)+\frac{2\alpha^2\beta}{1+\mu\alpha} +\frac{\alpha \sigma_i(1-p_i)}{2}\right)\Exp\left[\left\|\alpha_i^k-\nabla \phi_i({\bf A}_i^\top x^*)\right\|^2 \right] 
			\\&&\qquad -\sum_{i=1}^{n}\left( 2\gamma \lambda_i \alpha(1-\beta)- 2p_i^{-1}v_i \lambda_i^2\left(\alpha^2(1-\beta)+\frac{2\alpha^2\beta}{1+\mu\alpha}\right)-\alpha\sigma_i v_i \lambda_i^2 \right) \Exp\left[\left\| \nabla \phi_i({\bf A}_i^\top x)-\nabla \phi_i({\bf A}_i^\top x^*)\right\|^2\right]
		\end{eqnarray*}
		
		In order to guarantee that $\mathbb{E}[\Psi^{k+1}]\leq  \left( \frac{\beta}{1+\alpha\mu}+1-\beta\right)\mathbb{E}[\Psi^k]$, $\alpha$ and $\beta$ should be chosen such that 
		$$
		2p_i^{-1}v_i \lambda_i^2\left(\alpha^2(1-\beta)+\frac{2\alpha^2\beta}{1+\mu\alpha}\right)+\alpha\sigma_i v_i \lambda_i^2\leq 2\gamma \lambda_i \alpha(1-\beta),
		$$
		and 
		$$
		\alpha^2(1-\beta)+\frac{2\alpha^2\beta}{1+\mu\alpha} +\frac{\alpha \sigma_i(1-p_i)}{2} \leq  \frac{\alpha\sigma_i}{2} \left( \frac{\beta}{1+\alpha\mu}+1-\beta\right).
		$$
		for all $1\leq i\leq n$.  
		Now we let
		$
		\beta=\frac{1}{2}$ and $\delta\geq \alpha\left(\frac{1}{2}+\frac{1}{1+\mu\alpha}\right) $ so that $$1-\delta\mu\leq \frac{1}{2(1+\alpha\mu)}+\frac{1}{2}. $$ Then the above inequalities can be satisfied if
		\begin{align}\label{a:etd}
		p_i^{-1}v_i \lambda_i \delta+\frac{\sigma_i v_i\lambda_i}{2} \leq \frac{\gamma}{2},\quad \Rightarrow \quad \delta \leq \frac{\gamma p_i}{2v_i\lambda_i} - \frac{\sigma_i p_i}{2}, 
		\end{align}
		and
		\begin{align}\label{a:etd2}
		\delta+\frac{ \sigma_i(1-p_i)}{2} \leq \frac{\sigma_i}{2}(1-\delta \mu),\quad \Rightarrow \quad \delta \leq \frac{\sigma_ip_i}{\mu\sigma_i + 2},
		\end{align}
		for all $1\leq i\leq n$. Since $\sigma_i = \frac{\gamma}{2v_i\lambda_i}$, if $\delta$ satisfies
		$$
		\delta \leq \min_{1\leq i\leq n} \frac{p_i}{\mu + 4 v_i\lambda_i/\gamma},
		$$then~\eqref{a:etd} and~\eqref{a:etd2} hold.
		This means if we choose $\alpha$ such that
		$$
		\alpha\left(\frac{1}{2}+\frac{1}{1+\mu\alpha}\right) \leq  \min_{1\leq i\leq n} \frac{p_i}{\mu + 4 v_i\lambda_i/\gamma} ,
		$$
		then 
		$$\mathbb{E}[\Psi^{k+1}]\leq  \left( \frac{1}{2(1+\alpha\mu)}+\frac{1}{2}\right)\mathbb{E}[\Psi^k].$$
		In particular, we can choose
		$$
		\alpha\leq \frac{2}{3}\min_{1\leq i\leq n} \frac{p_i}{\mu + 4 v_i\lambda_i/\gamma}.
		$$
		Notice that when  we apply Lemma \ref{l:prox-svrg}, $\alpha$ should satisfy $\alpha \leq \frac{1}{L}$. Hence, 
		$$
		\alpha\leq \min\left\{\frac{2}{3}\min_{1\leq i\leq n} \frac{p_i}{\mu + 4 v_i\lambda_i/\gamma}, \frac{1}{L} \right\}.
		$$
		From $\mathbb{E}[\Psi^{k}]\leq  ( 1 - \frac{\alpha\mu}{2(1+\alpha\mu)})^k\mathbb{E}[\Psi^0]$, we know if 
		$$
		k \geq \left(2 + \max\left\{ \frac{2L}{\mu}, 3\max_{i} \left( \frac{1}{p_i} + \frac{4v_i\lambda_i}{\mu p_i\gamma} \right)  \right\} \right)\log\left(\frac{1}{\epsilon}\right)
		$$
		Then $\mathbb{E}[\Psi^{k}]\leq  \epsilon \mathbb{E}[\Psi^0]$.

\subsection{Proof of Theorem \ref{th:phistrongly}}	
	
		First notice that, if $\psi$ is $\mu$-strongly convex, then $P$ is $\mu$-strongly convex which implies that the optimal solution of problem (\ref{primal}) is unique. Let $ \cX^* = \{ x^*\}$. For the $x^*$, we have
		$$
		x^*=\prox^\psi_{\alpha}( x^*-\alpha \nabla f(x^*)).
		$$
		Therefore,
		\begin{eqnarray*}
			&&\mathbb{E}_k\left[ \left\|x^{k+1} - x^*\right\|^2 \right] \\ &=&
			\mathbb{E}_k\left[ \left\| \prox^\psi_{\alpha}(x^k  - \alpha g^k)-x^*\right\|^2 \right] \\
			&=&\mathbb{E}_k\left[ \left\| \prox^\psi_{\alpha}(x^k  - \alpha g^k)-\prox^\psi_{\alpha}(x^*-\alpha \nabla f(x^*)) \right\|^2\right]\\
			&\leq & \frac{1}{1+\alpha \mu}\mathbb{E}_k\left[ \left\| x^k  - \alpha g^k-(x^*-\alpha \nabla f(x^*)) \right\|^2\right]\\
			\qquad &\overset{(\ref{expgk2})}{=}& \frac{1}{1+\alpha \mu}\left\| x^k -x^*\right\|^2 - \frac{2\alpha}{1+\alpha \mu} \langle \nabla f(x^k)-\nabla f(x^*), x^k-x^* \rangle + \frac{\alpha^2}{1+\alpha \mu} \mathbb{E}_k\left[\| g^k-\nabla f(x^*) \|^2\right] \\
			&\overset{\text{Lemma~\ref{l:nadph} and \ref{lm:gk23}}}{\leq}& \frac{1}{1+\alpha \mu}\|x^k - x^*\|^2 + \frac{2\alpha}{1+\alpha \mu} \sum_{i=1}^{n} (\frac{\alpha v_i\lambda_i^2}{p_i} - \gamma \lambda_i)\left\| \nabla\phi_i({\bf A}_i^\top x^k)-\nabla \phi_i({\bf A}_i^\top x^*)\right\|^2 \\ 
			&& \qquad + \frac{2\alpha^2}{1+\alpha\mu}\sum_{i=1}^n p_i^{-1}v_i\lambda_i^2 \left\|\alpha^k_i - \nabla \phi_i({\bf A}_i^\top x^*)\right\|^2
		\end{eqnarray*}
	
	Taking expectation again and applying the tower property, we obtain  
	\begin{eqnarray*}
	\mathbb{E}\left[ \|x^{k+1} -x^*\|^2 \right] &\leq&  \frac{1}{1+\alpha \mu}\mathbb{E}[\|x^k - x^*\|^2] + \frac{2\alpha}{1+\alpha \mu} \mathbb{E}\left[\sum_{i=1}^{n} \left(\frac{\alpha v_i\lambda_i^2}{p_i} - \gamma \lambda_i\right)\left\| \nabla\phi_i({\bf A}_i^\top x^k)-\nabla \phi_i({\bf A}_i^\top x^*)\right\|^2\right] \\ 
	&& \qquad + \frac{2\alpha^2}{1+\alpha\mu}\mathbb{E} \left[\sum_{i=1}^n p_i^{-1}v_i\lambda_i^2 \left\|\alpha^k_i - \nabla \phi_i({\bf A}_i^\top x^*)\right\|^2\right]	
	\end{eqnarray*}

Therefore, for the stochastic Lyapunov function $\Psi^{k+1}$, we have 
\begin{eqnarray*}
	\mathbb{E}\left[ \Psi^{k+1}\right] &\overset{\text{Lemma}~\ref{lm:ak+1}}{\leq}& \frac{1}{1+\alpha \mu}\mathbb{E}[\|x^k-x^*\|^2] + \frac{\alpha}{1+\alpha \mu} \mathbb{E} \left[ \sum_{i=1}^n p_i^{-1}v_i\lambda_i^2 \left( 2\alpha + (1+\alpha \mu)\sigma_i(1-p_i)\right)\left\|\alpha^k_i - \nabla \phi_i({\bf A}_i^\top x^*)\right\|^2 \right] \\ 
	&&\qquad \frac{2\alpha }{1+\alpha \mu}\mathbb{E}\left[ \sum_{i=1}^n \lambda_i \left( \frac{\alpha v_i\lambda_i}{p_i} -\gamma + \frac{1+\alpha \mu}{2} \sigma_i v_i \lambda_i \right)\left\| \nabla\phi_i({\bf A}_i^\top x^k)-\nabla \phi_i({\bf A}_i^\top x^*)\right\|^2 \right]
\end{eqnarray*}
In order to guarantee that $\mathbb{E}[\Psi^{k+1}]\leq  \left( \frac{1}{1+\alpha \mu}\right)\mathbb{E}[\Psi^k]$, $\alpha$ should be choosen such that 
$$
\frac{\alpha v_i\lambda_i}{p_i} -\gamma + \frac{1+\alpha \mu}{2} \sigma_i v_i \lambda_i \leq 0, \quad \Rightarrow \quad \alpha \leq \frac{p_i\gamma}{v_i\lambda_i} - \frac{1+\alpha \mu}{2}p_i\sigma_i,
$$
and 
$$
2\alpha + (1+\alpha \mu)\sigma_i(1-p_i) \leq \sigma_i, \quad \Rightarrow \quad \alpha \leq \frac{\sigma_i p_i}{2+ \mu\sigma_i(1-p_i)},
$$
for all $1\leq i\leq n$. Assume $\alpha \leq 1/\mu$. Then the above inequalities can be satisfied if 
$$
\alpha \leq \frac{p_i\gamma}{v_i \lambda_i} - p_i \sigma_i, 
$$
and 
$$
\alpha \leq \frac{\sigma_i p_i}{2 + \mu \sigma_i},
$$
for all $i$. Since $\sigma_i = \frac{2\gamma}{3v_i \lambda_i}$, if $\alpha$ is chosen as 
$$
\alpha \leq \min_{1\leq i\leq n} \frac{p_i}{\mu + 3v_i\lambda_i/\gamma},
$$
which is actually satisfies $\alpha \leq 1/\mu$, then we have the recursion $\mathbb{E}[\Psi^{k+1}]\leq  \left( \frac{1}{1+\alpha \mu}\right)\mathbb{E}[\Psi^k]$.

	\newpage
	\section{Extra Experiments} \label{sec:more_experiments}
	
%\peter{Zheng, please add all experiments that could not fit into the main paper here.}	
We include in this section more experimental results.

\subsection{Batch sampling}\label{BS}
\begin{figure}[!h]
  \centering 
  \subfigure[SAGA; ijcnn1]{ 
    \label{fig:subfig:2a} %% label for first subfigure 
    \includegraphics[height=0.25\textwidth]{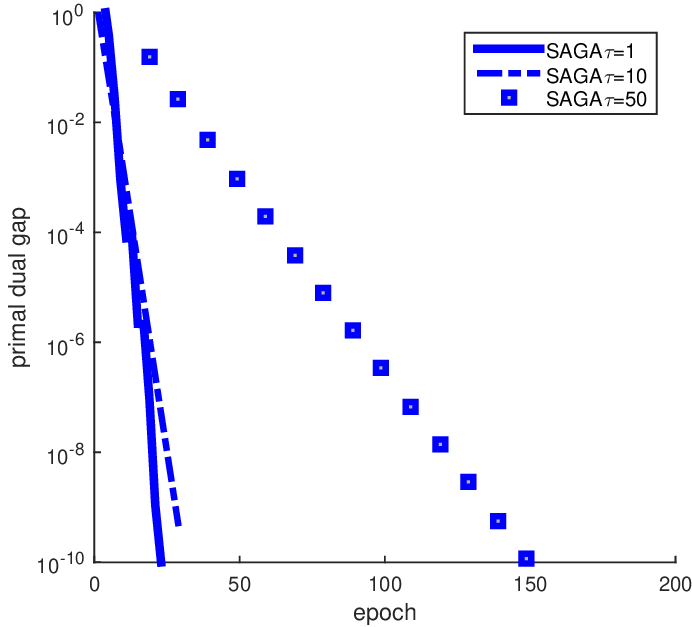}} 
  \subfigure[SDCA; ijcnn1]{ 
    \label{fig:subfig:2b} %% label for second subfigure 
    \includegraphics[height=0.25\textwidth]{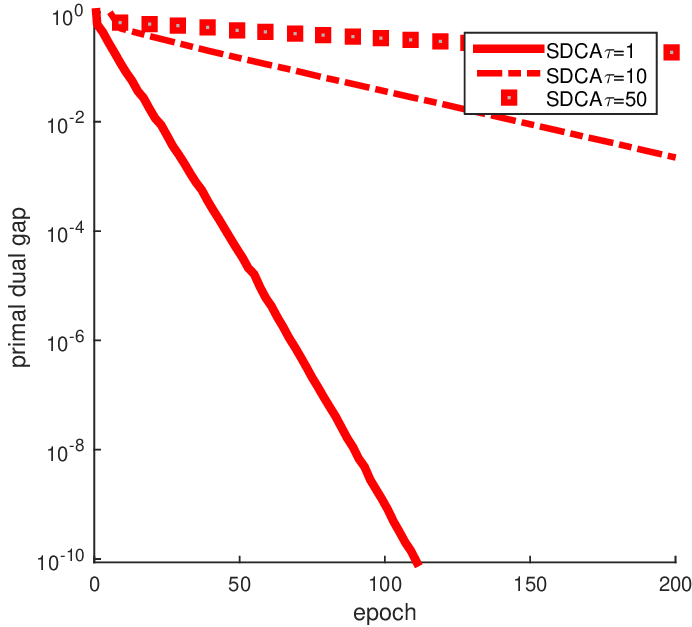}} 
  \caption{mini-batch SAGA V.S. mini-batch SDCA: ijcnn1}
  \label{figs1} %% label for entire figure 
\end{figure}

\begin{figure} [!h]
  \centering 
  \subfigure[SAGA; a9a]{ 
    \label{fig:subfig:3a} %% label for first subfigure 
    \includegraphics[height=0.25\textwidth]{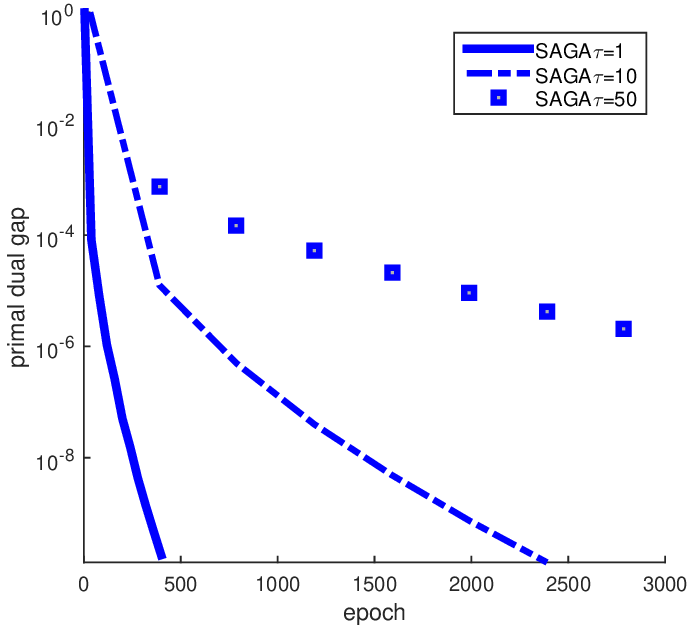}} 
  \subfigure[SDCA; a9a]{ 
    \label{fig:subfig:3b} %% label for second subfigure 
    \includegraphics[height=0.25\textwidth]{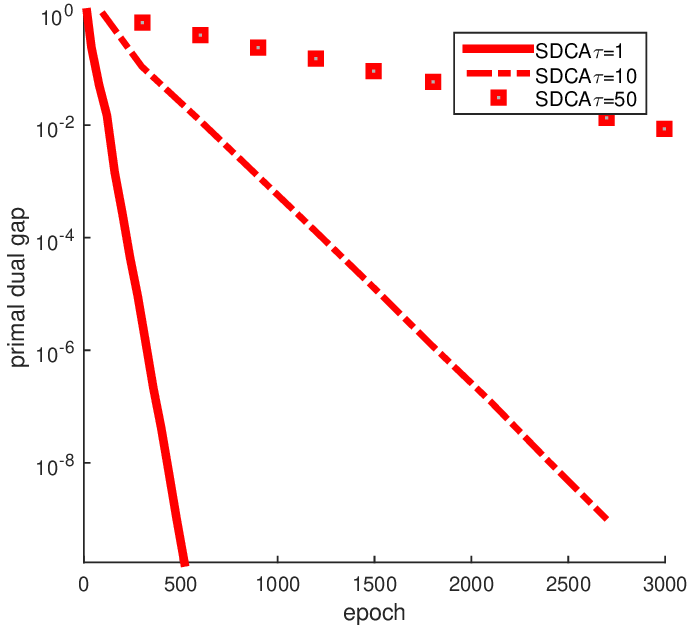}} 
  \caption{mini-batch SAGA V.S. mini-batch SDCA: a9a} 
  \label{figs3} %% label for entire figure 
\end{figure}

\begin{figure} [!h]
  \centering 
  \subfigure[SAGA; w8a]{ 
    \label{fig:subfig:4a} %% label for first subfigure 
    \includegraphics[height=0.25\textwidth]{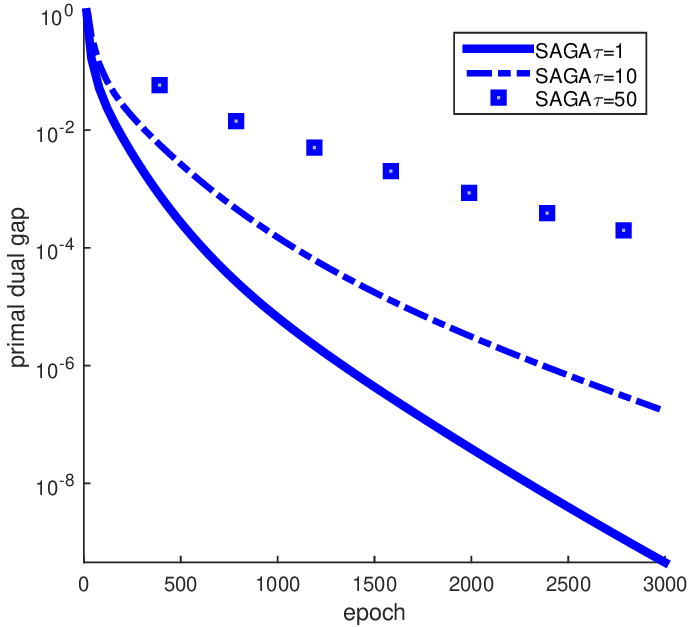}} 
  \subfigure[SDCA; w8a]{ 
    \label{fig:subfig:4b} %% label for second subfigure 
    \includegraphics[height=0.25\textwidth]{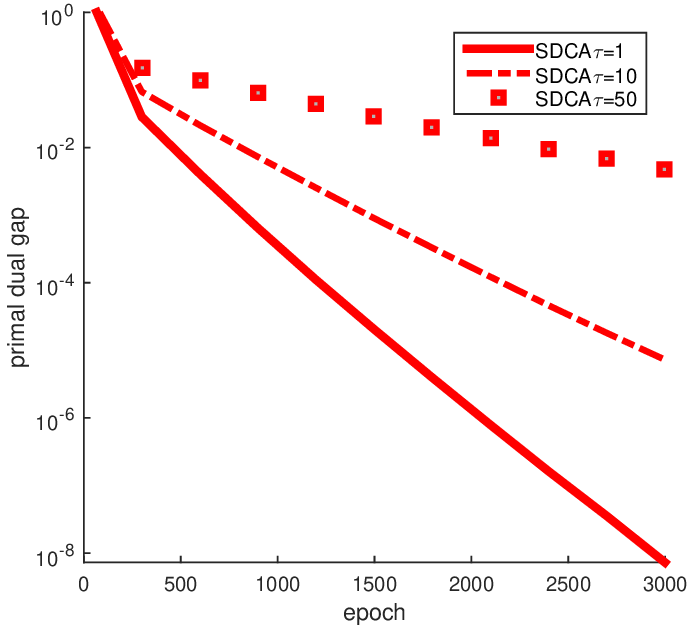}} 
  \caption{mini-batch SAGA V.S. mini-batch SDCA: w8a} 
  \label{figs4} %% label for entire figure 
\end{figure}

Here we compare SAGA-AS with  SDCA, for the case when $S$ is a $\tau$-nice sampling for three different values of $\tau\in \{1,10,50\}$. Note that SDCA with $\tau$-nice sampling works  the same both in theory and in practice as Quartz with $\tau$-nice sampling.  We report in Figure~\ref{figs1}, Figure~\ref{figs3} and Figure~\ref{figs4} the results obtained for the dataset ijcnn1, a9a and w8a.  Note that for ijcnn1, when we increase $\tau$ by 50, the number of epochs of SAGA-AS  only increased by less than 6. This indicates a considerable speedup if parallel computation can be included in the implementation of mini-batch case.

\subsection{Importance sampling}\label{IP}

\begin{figure} [!h]
  \centering 
  \subfigure[ijcnn1]{ 
    \label{fig:subfig:1a} %% label for first subfigure 
    \includegraphics[height=0.25\textwidth]{plots2/ijcnn1vs_IND.eps}} 
  \subfigure[w8a]{ 
    \label{fig:subfig:1b} %% label for second subfigure 
    \includegraphics[height=0.25\textwidth]{plots2/w8avs_IND.eps}} 
     \subfigure[a9a]{ 
    \label{fig:subfig:1c} %% label for second subfigure 
    \includegraphics[height=0.25\textwidth]{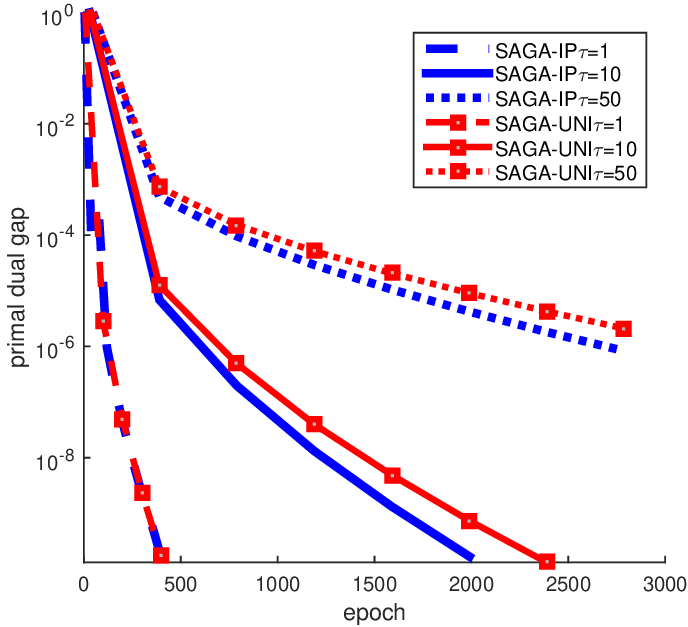}} 
  \caption{importance sampling V.S. uniform sampling} 
  \label{figs2} %% label for entire figure 
\end{figure}

We compare uniform sampling SAGA (SAGA-UNI) with importance sampling SAGA (SAGA-IP), as described in Section~\ref{sec:ip}
, on three values of $\tau\in\{1,10, 50\}$.   The results for the datasets w8a, ijcnn1  and a9a are shown in Figure~\ref{figs2}. 
For the dataset ijcnn1, mini-batch with importance sampling almost achieves linear speedup as the number of epochs does not increase with $\tau$. For the dataset w8a, mini-batch with importance sampling can even need less number of epochs than serial uniform sampling. For the dataset a9a, importance sampling slightly  but consistently improves over uniform sampling. Note that we adopt the importance sampling strategy described in~\cite{ACD2019} and the actual running time is the same as uniform sampling.

\subsection{Comparison with Coordinate Descent}\label{CD}

\begin{figure} [!h]
  \centering 
  \subfigure[ijcnn1]{ 
    \label{fig:subfig:5a} %% label for first subfigure 
    \includegraphics[height=0.25\textwidth]{plots2/ijcnn1vsCD.eps}} 
  \subfigure[w8a]{ 
    \label{fig:subfig:5b} %% label for second subfigure 
    \includegraphics[height=0.25\textwidth]{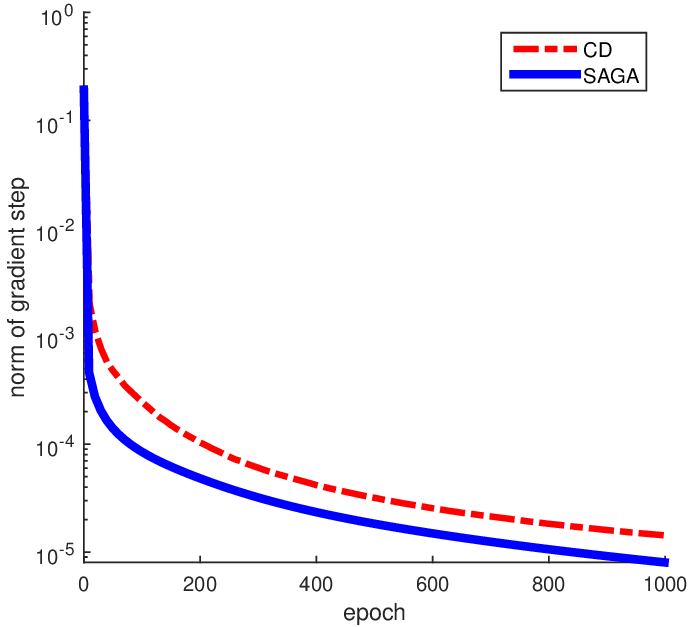}} 
     \subfigure[a9a]{ 
    \label{fig:subfig:5c} %% label for second subfigure 
    \includegraphics[height=0.25\textwidth]{plots2/a9avsCD.eps}} 
  \caption{SAGA V.S. CD} 
  \label{figs5} %% label for entire figure 
\end{figure}

We consider the un-regularized logistic regression problem~\eqref{a:log} with $\lambda_1=\lambda_2=0$. In this case, Theorem~\ref{th:phi} applies and we expect to have linear convergence of SAGA without any knowledge on the constant $\mu$ satsifying Assumption~\eqref{ass:rsc}, see Remark~\ref{rem:saga}. This makes SAGA comparable with descent methods such as gradient method and coordinate descent (CD) method. However, comparing with their deterministic counterparts, the speedup provided by CD can be at most of order $d$ while  the speedup by SAGA can be of order $n$.  Thus SAGA is much preferable  than CD when $n$ is larger than $d$.  We provide numerical evidence in Figure~\ref{figs5}.

\end{document}